\documentclass[lettersize,journal]{IEEEtran}
\usepackage{amsmath,amsfonts}
\usepackage{algorithmic}
\usepackage{algorithm}
\usepackage{array}
\usepackage[caption=false,font=normalsize,labelfont=sf,textfont=sf]{subfig}
\usepackage{textcomp}
\usepackage{stfloats}
\usepackage{url}
\usepackage{verbatim}
\usepackage{graphicx}
\usepackage{cite}
\usepackage{hyperref}
\usepackage{xcolor}
\usepackage{multirow}
\usepackage{lipsum}
\usepackage{orcidlink}
\begin{document}

\title{NeFT: Negative Feedback Training to Improve Robustness of Compute-In-Memory DNN Accelerators}

\author{
Yifan Qin~\orcidlink{0000-0001-6559-9913}, Zheyu Yan~\orcidlink{0000-0003-1830-606X}, Dailin Gan~\orcidlink{0000-0003-0535-1259}, Jun Xia,~\IEEEmembership{Member,~IEEE,} Zixuan Pan, Wujie Wen~\orcidlink{0000-0003-0011-0675},~\IEEEmembership{Member,~IEEE,} Xiaobo Sharon Hu~\orcidlink{0000-0002-6636-9738},~\IEEEmembership{Fellow,~IEEE,} 
Yiyu Shi~\orcidlink{0000-0002-6788-9823},~\IEEEmembership{Senior Member,~IEEE}
\thanks{Received 5 February 2025; revised 21 May 2025; accepted 17 July 2025. This article was recommended by Associate Editor M. Shafique.(\textit{Corresponding author: Yiyu Shi.})

Yifan Qin, Zheyu Yan, Jun Xia, Zixuan Pan, Xiaobo Sharon Hu,
and Yiyu Shi are with the Department of Computer Science and
Engineering, University of Notre Dame, Notre Dame, IN 46556 USA (e-mail: yqin3@nd.edu; shu@nd.edu; yshi4@nd.edu). 

Dailin Gan is with the Department of Applied and Computational
Mathematics and Statistics, University of Notre Dame, Notre Dame, IN 46556 USA.

Wujie Wen is with the Department of Computer Science, North Carolina State University, Raleigh, NC 27695 USA.

Digital Object Identifier 10.1109/TCAD.2025.3591409}
        }

\markboth{IEEE TRANSACTIONS ON COMPUTER-AIDED DESIGN OF INTEGRATED CIRCUITS AND SYSTEMS}%
{Yifan Qin \MakeLowercase{\textit{et al.}}: NeFT}

\IEEEpubid{
\parbox{1.7\columnwidth}{\centering
1937-4151~\copyright~2025~IEEE. All rights reserved, including rights for text and data mining, and training of artificial intelligence and similar technologies. Personal use is permitted, but republication/redistribution requires IEEE permission. \\See \url{https://www.ieee.org/publications/rights/index.html} for more information.}}

\maketitle

\begin{abstract}
Compute-in-memory accelerators built upon non-volatile memory devices excel in energy efficiency and latency when performing deep neural network (DNN) inference, thanks to their in-situ data processing capability. 
However, the stochastic nature and intrinsic variations of non-volatile memory devices often result in performance degradation during DNN inference.
Introducing these non-ideal device behaviors in DNN training enhances robustness, but drawbacks include limited accuracy improvement, reduced prediction confidence, and convergence issues. This arises from a mismatch between the deterministic training and non-deterministic device variations, as such training, though considering variations, relies solely on the model's final output.
In this work, inspired by control theory, we propose Negative Feedback Training (NeFT), a novel concept supported by theoretical analysis, to more effectively capture the multi-scale noisy information throughout the network. We instantiate this concept with two specific instances, oriented variational forward (OVF) and intermediate representation snapshot (IRS). Based on device variation models extracted from measured data, extensive experiments show that our NeFT outperforms existing state-of-the-art methods with up to a 45.08\% improvement in inference accuracy while reducing epistemic uncertainty, boosting output confidence, and improving convergence probability. These results underline the generality and practicality of our NeFT framework for increasing the robustness of DNNs against device variations.
The source code for these two instances is available at \href{https://github.com/YifanQin-ND/NeFT_CIM}{https://github.com/YifanQin-ND/NeFT\_CIM}.
\end{abstract}


\section{Introduction}
\IEEEpubidadjcol
\IEEEPARstart{D}{eep} neural networks (DNNs) have profoundly transformed numerous domains in modern society, as exemplified by breakthroughs like transformers~\cite{vaswani2017attention} and stable diffusion~\cite{rombach2022high}. Despite these advances, accelerating DNN inference—which demands intensive vector-matrix operations—remains constrained by frequent data transfers between memory and processing units. Historically, in developing the first programmable computer, the EDVAC~\cite{von1993first}, heterogeneous technologies for computation and memory were employed separately, driven by disparate speeds and costs. Over time, this led to the establishment of a memory hierarchy, which continues to shape modern computing architectures. As energy and bandwidth consumption for off-chip memory dominate, the resulting “memory wall”~\cite{mckee2004reflections} further exacerbates the von Neumann bottleneck.
\IEEEpubidadjcol
This bottleneck is particularly pronounced in neural inference, where the large-scale vector-matrix multiplication requires substantial data movement. Conventional DNN accelerators such as GPUs often face limitations in latency, energy, privacy, and sustainability due to data-intensive workloads. A promising solution involves non-volatile memory-based computing-in-memory (NVCIM) DNN accelerators, which store network weights and perform in-situ vector-matrix multiplication within the same crossbar array in $O(1)$ time by leveraging Kirchhoff’s current law. Notably, emerging NVM devices offer improved memory density and energy efficiency~\cite{chen2016eyeriss}.

Despite these advantages, NVM devices inherently suffer from non-idealities like cycle-to-cycle and device-to-device variations~\cite{shim2020two}, often caused by thermal fluctuations, radiation, and fabrication defects. Such variations induce deviations in device conductance values after programming~\cite{raty2015aging, diware2023mapping}, degrading the precision of stored model weights and undermining inference accuracy~\cite{yan2021uncertainty}. Furthermore, factors such as read disturbance, lattice relaxation, and defect accumulation~\cite{rizzi2011role, raty2015aging, chen2009highly, lee2010comprehensively} exacerbate conductance drift over time. Consequently, a Gaussian-distributed deviation in programmed conductance often arises, leading to performance degradation in NVCIM DNN accelerators~\cite{yan2021uncertainty, wang2022device}.
\begin{figure}[t]
  \centering
  \includegraphics[scale=0.32]{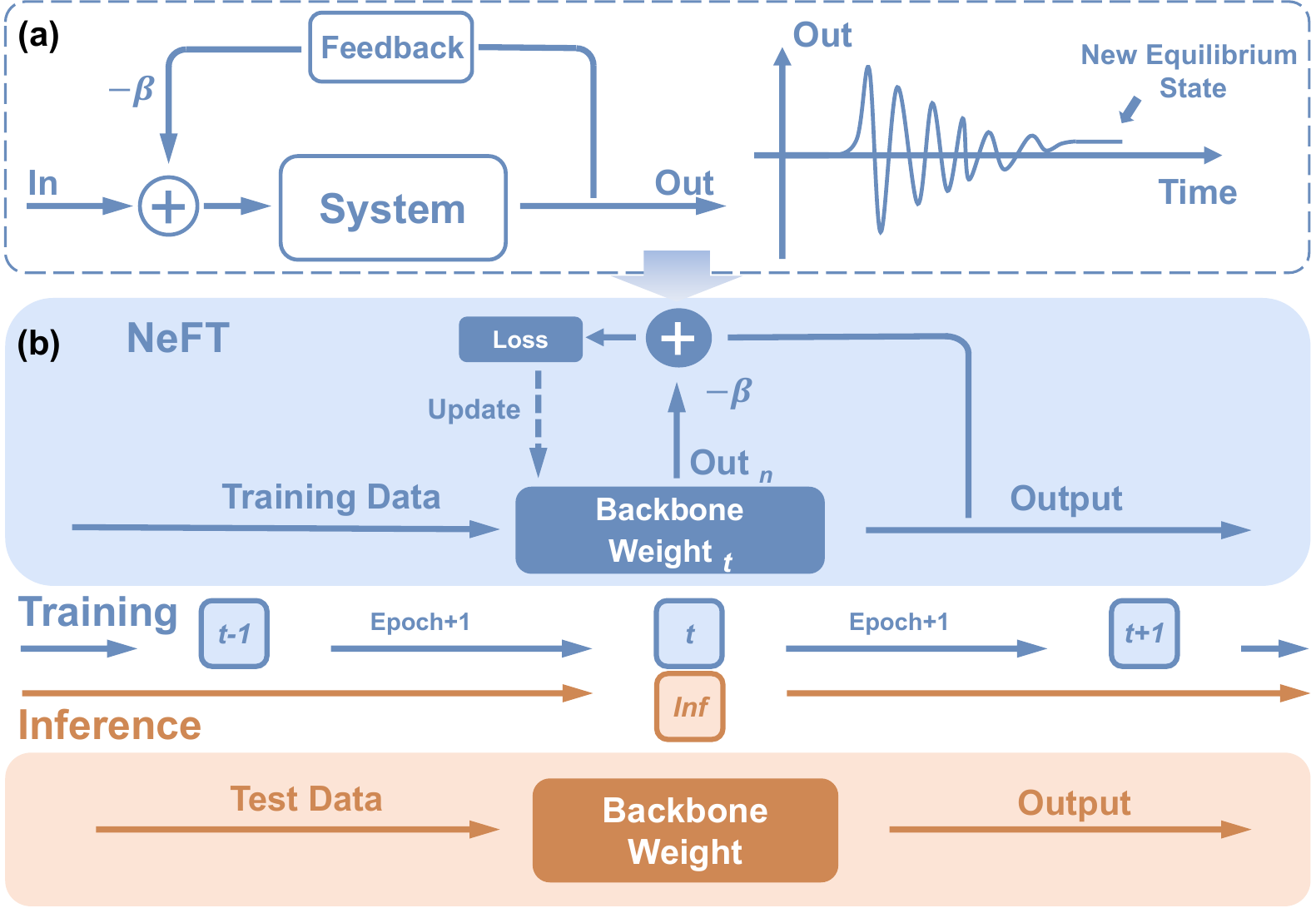}
  \caption{(a) A schematic illustration of a classic negative feedback system and its transition to a new equilibrium. (b) A conceptual diagram of the proposed negative feedback training method.}
  \label{fig: overview}
  \vspace{-0.2in}
\end{figure}

\IEEEpubidadjcol
Achieving reliable DNN inference on variational NVM devices is challenging.
A variety of hardware and software solutions have been proposed, including 
device upgrades~\cite{degraeve2015causes}, write-verify~\cite{shim2020two}, 
variation-aware training~\cite{eldebiky2023correctnet, zhu2020statistical, jiang2020device, qin2024tsb}, and tiny shared block  techniques~\cite{qin2024tsb}. Among these, variation-aware training~\cite{jiang2020device, yang2022tolerating, yan2023improving, peng2019dnn+} has gained popularity due to its effectiveness and its ability to function without hardware modifications to the accelerator. By simply exposing DNNs to the target noise (i.e., modeling device variations)~\cite{qin2020design, yan2022computing, wan2022accuracy} during training, the network’s tolerance to noise is improved.

However, state-of-the-art (SOTA) variation-aware training methods still face challenges, including limited accuracy improvements, reduced prediction confidence (often evidenced by increased epistemic uncertainty~\cite{gawlikowski2023survey}), and difficulties in achieving convergence under large variations. Our analysis
indicates that these shortcomings are mainly due to a mismatch between the nondeterministic nature of noise and the deterministic training framework, for two main reasons: \textbf{First}, a finite number of training epochs can only expose the network to a limited set of noise samples, limiting the model’s ability to fully learn the characteristics of the noise. \textbf{Second}, compared with standard training, variation-aware training can yield more diverse optimization directions, some of which may lead to suboptimal or incorrect states. The greater uncertainty of the output exacerbates this problem. Consequently, the network may not converge to an optimal solution, and even a seemingly well-trained model may produce low-quality predictions when faced with device variations during inference.

We hypothesize that the aforementioned mismatch can be mitigated by allowing the model to learn sufficient variation information from multiple components throughout the training process, rather than relying solely on the final output, as is common in existing SOTA methods. This conjecture is grounded in modern control
theory, which highlights the pivotal role of negative feedback in enhancing a system’s stability, reducing steady-state error, and mitigating external disturbances. When a balanced system experiences noise or perturbations, negative feedback helps it resist such disturbances and eventually settles into a new equilibrium state,
as depicted in Figure~\ref{fig: overview}(a). In this setting, a portion of the outputs generate feedback signals, whose strength is governed by the negative feedback coefficient~$\beta$. Adjusting $\beta$, the system can dynamically moderate the influence of disturbances to maintain robust operation.

Drawing inspiration from these control-theoretic principles, we introduce \textbf{Ne}gative \textbf{F}eedback \textbf{T}training (\textbf{NeFT}) for DNNs. In contrast to conventional variation-aware training that focuses predominantly on the final output, NeFT takes advantage of negative contributions from multiple parts of the network’s intermediate and final outputs to mitigate noise impacts, thereby guiding DNN to a more robust state. At a high level, the entire NeFT process can be described as follows: During training, NeFT augments the DNN backbone with additional negative feedback components that capture variation-related signals from the model. These negative feedback signals ($Out_n$), scaled by $\beta$, have negative impact during the weight-update process, effectively embedding multi-scale noise information into the model’s learning dynamics. Once the training phase completes, all negative feedback components are removed, leaving an unaltered, yet more robust, DNN backbone (see Figure~\ref{fig: overview}(b)). Importantly, the DNN architecture itself remains unchanged during inference, ensuring no additional hardware or latency overhead. To better illustrate the analogy with a classical negative feedback system, we clarify the system input-output structure of NeFT in each training iteration. Specifically, the system input consists of the model weights and training data, while the output includes the predicted labels and the updated model weights. A portion of the output—namely, feedback signal—is fed back to the training loss to constrain weight updates. This feedback mechanism guides the model towards a new equilibrium state with improved robustness against device variation.

Within this framework, the notion of “negative” signifies diminishing the detrimental effects of the target variations while still preserving their essential informational content, whereas the term “feedback” emphasizes the introduction of supplementary output signals distinct from those available in the backbone’s direct outputs. When combined, this \textit{negative feedback} mechanism constrains network optimization based on the noise itself, enforcing stronger corrections whenever perturbations are greater. As a result, the network remains closer to an optimal solution path, which facilitates stable convergence even under significant variations in device conductance. This approach departs from existing strategies that rely solely on final-layer outputs, allowing NeFT to capture nuanced, intermediate variation patterns that might otherwise go unnoticed. Through this multi-scale lens, NeFT demonstrates improved tolerance to large variations, thereby achieving higher inference accuracy and reliability in practical NVM-based compute-in-memory accelerators.

To this end, we present two implementation instances of NeFT to demonstrate its broad applicability and practicality for enhancing the robustness of DNN accelerators under device variations. These two instances are named oriented variational forward (OVF) and intermediate representation snapshot (IRS). While OVF optimizes the network from the perspective of overall variational performance---emphasizing stable inference across multiple variation scenarios---IRS focuses on internal representations within the network, capturing and constraining feature-level signals during training. Both instances employ different designs and ideas but share the negative feedback training concept.

Our main contributions are as follows:
\begin{itemize}
    \item We introduce a negative feedback mechanism into the DNN training process to stabilize it and enhance the network’s robustness against device variations. \textit{To the best of our knowledge, this is the first negative feedback-based training method aimed at improving robustness in NVCIM accelerators.}
    \item We propose two novel implementations of negative feedback training—namely, oriented variational forward (OVF) and intermediate representation snapshot (IRS)—which address device variations from an overall variational performance perspective and through internal feature representations, respectively.
    \item We provide a theoretical analysis that supports the effectiveness and stability of our proposed NeFT framework, further validating its underlying rationale.
    \item Our extensive simulations demonstrate NeFT’s effectiveness in boosting accuracy, enhancing confidence, and improving convergence probability. Notably, NeFT achieves up to 45.08\% improvement in DNN average inference performance compared to SOTA methods.
\end{itemize}

We believe this work represents an important step in bridging the gap between deterministic 
training and non-deterministic device variations in DNN models for CIM accelerators, 
highlighting NeFT’s potential as a promising direction for improving robustness. 
The remainder of this paper is organized as follows: 
Section~\ref{sec:background} introduces the necessary background and reviews 
state-of-the-art solutions for enhancing the robustness of NVCIM accelerators. 
Section~\ref{sec:proposed method} presents the fundamental principles of our 
NeFT method and provides detailed descriptions of the two proposed implementations, 
OVF and IRS. Section~\ref{sec:experiments} demonstrates the effectiveness of NeFT 
through comprehensive experiments, and finally, Section~\ref{sec:conclusion} 
concludes the paper.
\vspace{-0.2in}



\section{Background and Related works}\label{sec:background}

In this section, we first introduce the fundamental structure of CIM DNN accelerators, then discuss the robustness challenges they face due to device variations,
and finally emphasize existing efforts aimed at addressing these issues.
\vspace{-0.2in}

\subsection{Computing-in-memory Crossbar}
CIM offers an innovative solution to address the limitations of conventional von Neumann architectures by tightly integrating computation and memory within the same memory array (or crossbar)~\cite{shafiee2016isaac}.
In such crossbar structures, matrix values (e.g., DNN weights) are stored at the intersection of vertical and horizontal lines using NVM devices, such as ferroelectric field-effect transistors (FeFETs)~\cite{reis2018computing}, resistive random-access memories (RRAMs)~\cite{qin2020design}, magnetoresistive random-access memories (MRAMs)~\cite{angizi2019mrima}, and phase-change memories (PCMs)~\cite{sun2021pcm}.
By applying carefully designed voltage pulses to specific rows, the crossbar can accumulate currents following Kirchhoff's current laws, thereby completing matrix multiply-accumulate (MAC) operations and certain non-linear transformations in a single clock cycle.
This approach effectively eliminates the need for data movement between processing and memory units.
In addition to matrix MAC operations, other crucial DNN operations (e.g., pooling and non-linear activation) are executed by peripheral digital circuits.
Consequently, digital-to-analog and analog-to-digital converters (DACs/ADCs) serve to bridge the analog and digital domains, ensuring the full functionality of CIM systems.
\vspace{-0.2in}

\subsection{Robustness against Device Variation}

NVM devices are susceptible to various sources of variation and noise, primarily including spatial and temporal variations, which can degrade inference accuracy after weight programming.
Spatial variations arise from fabrication defects and may exhibit both localized and widespread correlations across devices. In contrast, temporal variations result from stochastic fluctuations in device materials and conductive mechanisms. Unlike spatial variations, temporal variations are typically independent of specific devices yet depend on the programmed values~\cite{feinberg2018making}.
For instance, the same NVM device may exhibit different conductance values even under identical programming pulses, and the amplitude of these variations can differ for various target conductance levels.

In this work, we assume that non-idealities are uncorrelated across distinct NVM devices 
but correlated with their programmed values. Specifically, we adopt a Gaussian-based abstract variation model~\cite{yan2022computing, yan2023improving, qin2020design, qin2024tsb, doevenspeck2021oxrram} and simulate four different device models to evaluate the general effectiveness of the NeFT method and its two instances. Our approach can be further adapted to other distributions of variations through appropriate modifications to the modeling assumptions.
\vspace{-0.2in}

\subsection{Existing Works and Evaluation Method}

To address the DNN performance degradation caused by the aforementioned NVM device variations,
two main research directions have emerged:
1) reducing device conductance variation at the hardware level, and
2) improving DNN robustness in the presence of device variations at the software level.

On the hardware side, device upgrades can be achieved through advancements in
material technology and fabrication processes~\cite{degraeve2015causes}.
Additionally, a commonly adopted approach involves write-verify~\cite{shim2020two} operations during device programming. In this method, the device is programmed with a predefined pulse, followed by a reading pulse to verify whether the resulting conductance falls within the target range. If not, further write-verify pulses are iteratively applied until the conductance value meets the desired specification. Although effective in reducing conductance deviations, this procedure often
becomes time-consuming due to repeated programming and verification steps. Recent research indicates that selectively applying write-verify only to critical devices can maintain accuracy while mitigating overhead~\cite{yan2022swim}.

On the software side, variation-aware training~\cite{eldebiky2023correctnet, zhu2020statistical, jiang2020device, qin2024tsb, wang2022device} has proven to be an effective strategy for enhancing DNN robustness to device variations. Generally, three key approaches fall under this category: regularization-based methods, statistical training, and noise-aware training. For instance, regularization-based methods introduce specific terms into the training objective to improve the network’s resilience. As one example,
CorrectNet~\cite{eldebiky2023correctnet} employs a modified Lipschitz constant regularization to enhance the tolerance of DNN weights to device variations. Statistical-based methods~\cite{zhu2020statistical} treat variations and noise as correlated random variables, incorporating them into the objective function.
Noise-aware training~\cite{jiang2020device, qin2024tsb} is among the most prominent software-based solutions, injecting variation into weights during training so that the model learns to cope with potential deviations. In each training iteration, a variation sample is drawn from the target noise distribution and added to the weights in the forward pass. The unperturbed weights are then updated by gradients influenced by this variation sample.

To evaluate the robustness of CIM accelerators against device variations, researchers often employ Monte Carlo (MC) simulations. Based on physical measurements, device and circuit models are established, and the target DNN model is mapped onto this circuit representation, initializing the NVM devices with desired conductance values. For each MC run, a non-ideal state is applied based on the variation model, assigning actual NVM conductance values that deviate from the ideal. Key metrics such as inference accuracy are then collected over multiple MC runs—often in the thousands~\cite{yan2022swim}—to obtain statistically meaningful results.
\vspace{-0.2in}
\section{Proposed method}\label{sec:proposed method}

In this section, we incorporate the negative feedback mechanism into the training loop and introduce a novel approach called negative feedback training with theoretical analysis, specifically designed to improve the robustness of NVCIM DNN accelerators under device variations. By embedding multi-scale variation signals into the optimization process, NeFT effectively guides network weights toward more stable and noise-resilient configurations. Building on the NeFT framework, we propose two distinct implementations: oriented variational forward (OVF) and intermediate representation snapshot (IRS). While OVF optimizes the network based on overall variational performance, IRS focuses on internal feature representations during the training process. Although each instance adopts a different strategy, both adhere to the core principle of negative feedback, offering flexible solutions for mitigating device variations.
\vspace{-0.2in}

\subsection{Negative Feedback Training (NeFT)}\label{sec: NeFT method}

Our proposed method draws inspiration from negative feedback theory, a fundamental principle in control systems. According to this theory, when a system is subjected to perturbations or disturbances, it utilizes feedback from its output to suppress or attenuate these disturbances, ultimately settling into a new equilibrium state via self-adjustment. In our context, the weight noise induced by device variations is treated as a form of perturbation affecting the DNN system. By incorporating negative feedback, we aim to improve system robustness and diminish the detrimental effects of such ``noise'' on the network’s performance. NeFT consists of two main components: a backbone DNN architecture and a negative feedback loop. During training, the feedback loop stabilizes the optimization process and guides the weights toward more noise-resilient configurations. Once training is complete, all feedback components are removed for inference, ensuring minimal overhead in practical deployments. Naively utilizing a negatively scaled output of DNN as a feedback signal, as in standard negative feedback systems, is not viable because it only leads to a direct scaling of the loss function without modifying the training method.

\begin{figure}[t]
  \centering
  \includegraphics[scale=0.35]{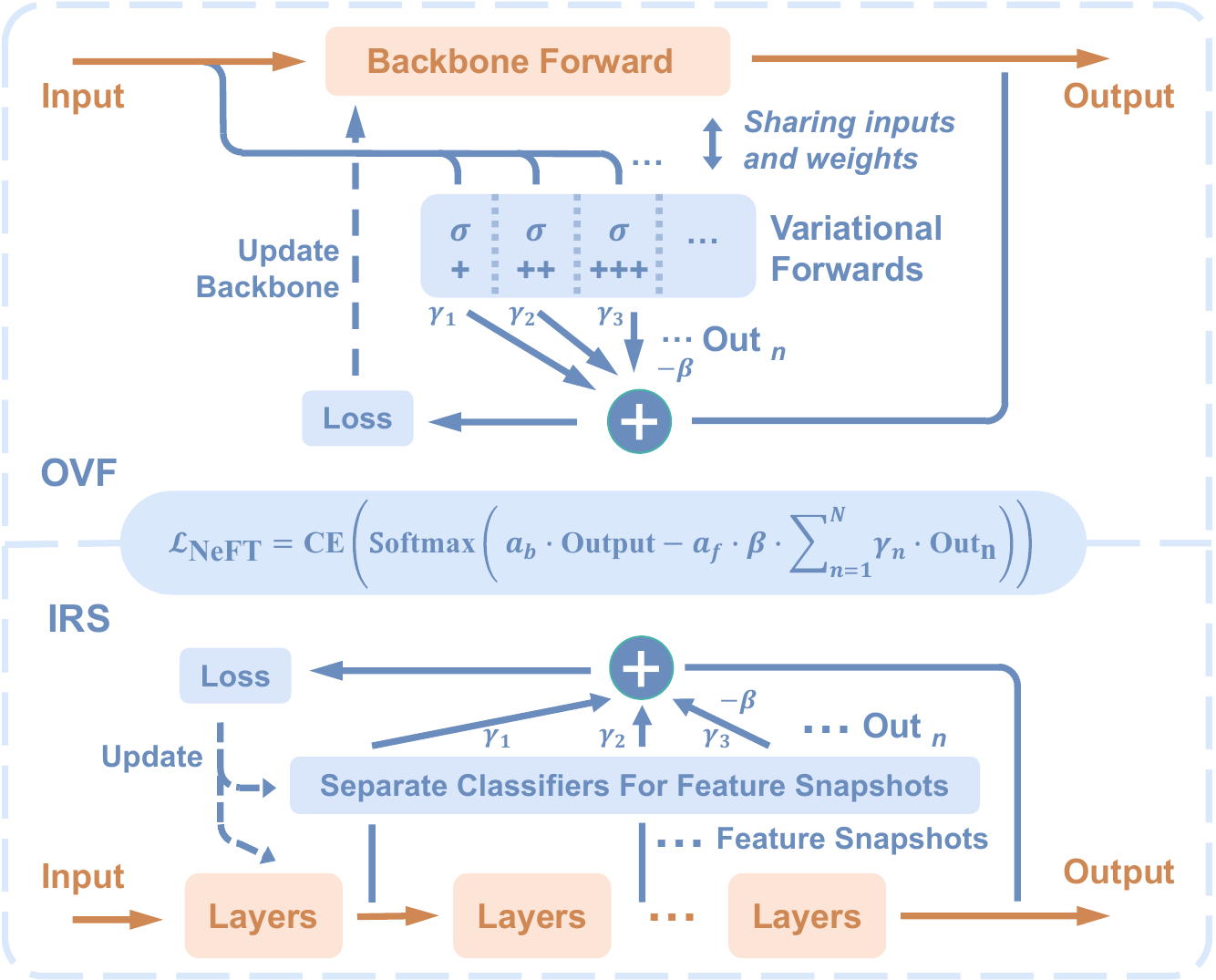}
  \caption{Two example NeFT implementations: oriented variational forward (upper) and intermediate representation snapshot (lower).}
  \label{fig: two instances images}
  \vspace{-0.2in}
\end{figure}

Instead, we require a negative feedback loop that can track changes in the output while being relatively distinct from it. This ensures that the feedback signals can truly reflect the impact of noise on the DNN weights and outputs while providing new angles to optimize the weight robustness.
To meet this requirement, negative feedback must satisfy two criteria. First, it should be generated by the components that are influenced by the same noise pattern present in the backbone.
Second, the negative feedback should have a strong connection with the backbone weights, ensuring that it accurately reflects weight perturbations within the backbone. In NeFT, we employ a distinct transformation of the outputs as feedback signals. Aiming for accurate predictions, the feedback deviating further from the target should have a larger constraint during training. In a DNN system with parameters $\mathbf{W} \in \mathbb{R}^d$, let $\mathbf{x}$ be the network input and $\hat{\mathbf{y}}$ the desired (one-hot) label, and variations arise as a random perturbation $\boldsymbol{\Delta}$. The DNN output can be written as $\mathcal{O}_\text{backbone}(\mathbf{x};\mathbf{W}+\boldsymbol{\Delta}) := \mathrm{Output}$. Concretely, let $\{\,Out_{n}(\mathbf{x}; \mathbf{W}+\boldsymbol{\Delta})\}_{n=1}^N$ be a set of $N$ auxiliary feedback outputs capturing noise-related information. Each $Out_{n}$ is scaled by $\gamma_n$ and summed to form a composite feedback term
\vspace{-0.1in}
\begin{equation}\label{eq:feedback output}
    \mathcal{O}_\text{feedback}(\mathbf{x};\mathbf{W}+\boldsymbol{\Delta}):=\; \sum_{n=1}^{N} \gamma_n \, Out_{n}
\end{equation}
By introducing a negative feedback coefficient $\beta$, we define the total output as:
\vspace{-0.1in}
\begin{equation}\label{eq:total output}
    \mathcal{O}_\text{total}(\mathbf{x};\mathbf{W}+\boldsymbol{\Delta}) \;:=\; a_{b}\,\mathcal{O}_\text{backbone}-a_{f}\,\beta\,\mathcal{O}_\text{feedback}
\end{equation}
where $a_{b}$ and $a_{f}$ are scaling factors regulating the respective contributions of the backbone and feedback signals. We then adopt the cross-entropy criterion, 
\vspace{-0.1in}
\begin{equation}\label{eq:loss}
    \mathcal{L}_\text{NeFT}(\mathbf{x}, \hat{\mathbf{y}}) 
    \;=\; -\,\langle \hat{\mathbf{y}}, \log\!\bigl(\mathrm{softmax}(\mathcal{O}_\text{total}(\mathbf{x}))\bigr)\rangle
\end{equation}
as our training objective. During each iteration, we sample $\boldsymbol{\Delta}$ from a device-specific distribution to simulate NVM variability and then compute the gradient $\nabla_{\mathbf{W}}\mathcal{L}_\text{NeFT}$ in a deterministic manner. Importantly, only the \emph{variation-free} parameters $\mathbf{W}$ are updated, effectively steering the system toward solutions that exhibit reduced sensitivity to device variation. Once training completes, we remove all negative feedback structures, leaving a standard DNN with parameter set $\mathbf{W}$. In this way, negative feedback serves to suppress perturbations arising from device variations, guiding $\mathbf{W}$ toward a noise-resilient equilibrium analogous to equilibrium points in classical control systems. Various schemes can be employed to instantiate $Out_{n}$, two of which—oriented variational forward and intermediate representation snapshot—are depicted in Figure~\ref{fig: two instances images}, highlighting distinct design angles for capturing and mitigating noise effects within the network.

Compared with existing variation-aware training methods, NeFT introduces a fundamentally different perspective. Conventional approaches such as noise-injection training (W/ Noise) expose the model to random perturbations during training but lack mechanisms to control or guide the learning path. Other methods, such as CorrectNet, apply output-level regularization (e.g., Lipschitz constraints) but only regulate the final prediction, ignoring the evolution of internal representations.
In contrast, NeFT imposes feedback constraints derived from intermediate outputs—such as feature activations or variational forward signals—which dynamically influence the direction of parameter updates. This feedback-driven training mechanism enables NeFT to stabilize learning and enhance robustness without requiring manually control.
The principle behind NeFT—that a system can regulate itself through internal feedback—offers a novel and interpretable approach to bridging the gap between non-deterministic hardware behavior and deterministic training processes. It also exemplifies how classical control theory can inform and enhance modern deep learning methods.

\vspace{-0.2in}

\subsection{Oriented Variational Forward (OVF)}
\label{sec:OVF method}

We present a specific implementation called oriented variational forward (OVF), depicted in the upper portion of Figure~\ref{fig: two instances images}. OVF uses the same backbone weights $\mathbf{W}$ but generates multiple feedback outputs by running forward passes with noise levels exceeding the standard device-specific variations. By feeding these outputs back into the NeFT framework, OVF effectively constrains the backbone from drifting away from the optimal optimization direction.

During each training iteration $i$, we first sample a noise instance $\Delta\mathbf{W}_i \sim \mathcal{N}(0,\sigma^2)$ to reflect typical device variations and obtain the primary backbone output $\mathcal{O}_\text{backbone}$ using $\mathbf{W}+\Delta\mathbf{W}_i$. Subsequently, we perform multiple oriented variational forwards with larger noise levels $\sigma$, each producing a feedback output $Out$. Once these N feedback outputs have been collected, we compute the negative feedback signal and apply it to the NeFT objective. The entire process is summarized in Algorithm~\ref{OVF algo}.

Consider the weight noise $\Delta\mathbf{W}$ raised by device variation. In a purely noise-injection scheme without feedback, the parameter update may deviate significantly in response to the perturbations. By contrast, OVF augments the objective with multiple ``oriented'' noisy forwards, each having a higher noise variance $\sigma + \Delta\sigma$. These oriented forwards effectively penalize parameter configurations that are overly sensitive to variations, thereby guiding the weights $\mathbf{W}$ toward a region where $\Delta\mathbf{W}$ have reduced impact on the final accuracy.

More concretely, one can interpret each oriented forward as probing the local Lipschitz smoothness of the model. If a particular set of weights $\mathbf{W}$ yields significantly different outputs under slightly amplified noise, the negative constraint from those forward passes will be larger, pushing the model to adjust toward a flatter—and thus more robust—region of the loss landscape. Such a mechanism aligns with the robustness arguments in adversarial or noise-augmented training, indicating that solutions lying in ``flatter minima'' are less susceptible to performance degradation caused by device variation. Empirically, we observe improved convergence behavior and higher inference accuracy in NVM variations.

In practice, increasing the standard deviation $\sigma$ injects more entropy into the system, leading to larger deviations from the target for the oriented forwards. Consequently, the feedback outputs associated with higher $\sigma$ exert a stronger negative influence on the backbone, enforcing stricter constraints. Mathematically, we assign decay factors $\gamma_{n}=10^{n-N}$ to highlight the negative feedback on the backbone model. All these oriented variational forwards share the same variation-free backbone $\mathbf{W}$ to ensure a common reference point, and maintain consistent Gaussian noise patterns, finally satisfying the criteria in Section~\ref{sec: NeFT method}.

\begin{algorithm}[t]
    \renewcommand{\algorithmicrequire}{\textbf{Input:}}
	\renewcommand{\algorithmicensure}{\textbf{Output:}}
    \caption{OVF($\mathcal{M}$, $\mathbf{W}$, $ep$, $\mathbf{D}$, $\alpha$, $\beta$, $N$)}
    \label{OVF algo}
    \begin{algorithmic}[1]
    \REQUIRE DNN backbone topology $\mathcal{M}$, weight $\mathbf{W}$, number of epochs $ep$, dataset $\mathbf{D}$, learning rate $\alpha$, negative feedback coefficient $\beta$, number of oriented forwards $N$;
    \FOR{($i=0$; $i < ep$; $i++$)}
        \FOR{$(x, \hat{y})$ in $\mathbf{D}$}
            \STATE Sample $\Delta\mathbf{W}_i \sim \mathcal{N}(0,\sigma^2)$;
            \STATE $\mathcal{O}_\text{backbone} \leftarrow \mathcal{M}(\mathbf{W}+\Delta\mathbf{W}_i, x)$;
            \FOR{($n=1$; $n \le N$; $n++$)}
                \STATE $\sigma \leftarrow \sigma + \Delta\sigma$; 
                \STATE Sample $\Delta\mathbf{W}_n' \sim \mathcal{N}(0,\sigma^2)$;
                \STATE $Out_n \leftarrow \mathcal{M}(\mathbf{W}+\Delta\mathbf{W}_n', x)$;
            \ENDFOR
            \STATE Restore $\sigma$;
            \STATE $\mathcal{O}_\text{feedback} \leftarrow \sum_{n=1}^{N} 10^{n-N} \cdot Out_n$;
            \STATE $\mathcal{O}_\text{total} \leftarrow a_{b}\,\mathcal{O}_\text{backbone} - a_{f}\,\beta\,\mathcal{O}_\text{feedback}$;
            \STATE $loss(x,\hat{y}) \leftarrow \text{CrossEntropy}(\mathcal{O}_\text{total}, \hat{y})$;
            \STATE Update $\mathbf{W}$ by gradient descent on $loss$, using the variation-free weight reference.
        \ENDFOR
    \ENDFOR
    \ENSURE Trained, robust DNN backbone $\mathcal{M}(\mathbf{W})$.
    \end{algorithmic}
\end{algorithm}

\subsection{Intermediate Representation Snapshot (IRS)}
\label{sec: IRS method}

After introducing NeFT, we present another specific implementation called intermediate representation snapshot (IRS). As illustrated in the lower part of Figure~\ref{fig: two instances images}, IRS leverages internal feature representation snapshots as negative feedback during training, enabling the network to observe and regulate data representations at multiple depths. By applying negative feedback to these intermediate features, IRS ensures that the noise-induced perturbation within internal layers is reflected in the overall objective and thus mitigated.

IRS adds transformation modules corresponding to different feature representations in the DNN, each acting as an independent probabilistic classifier. These modules map intermediate features to the same output dimension as the backbone’s final classifier, maintaining a consistent probability distribution shape. Notably, each $Out_n$ reuses a portion of the backbone’s weights and operations, tying the snapshot’s performance closely to that of the backbone network.

Selecting where and how many snapshots to place often follows various heuristic or domain-specific principles. In our approach, we apply the \emph{semantics principle}: convolution layers with the same kernel depth form one block, and we introduce a snapshot at the end of each block. To reduce confounding factors related to different sizes of classifiers, all snapshot classifiers share nearly identical shapes.

In classical DNN, relying solely on the system’s final output (akin to the backbone’s last-layer output) may delay or obscure disturbances occurring within intermediate states. By contrast, the IRS provides real-time access to these internal “states” (i.e., intermediate feature maps) through distinct snapshot classifiers. This multi-point sensing of the network’s internal feature space improves disturbance rejection. If the intermediate representations are highly sensitive to a small noise in $\mathbf{W}$ (caused by device variation), the negative feedback derived from the target will become larger, pushing the weight update to favor a more robust area. Furthermore, the IRS effectively penalizes local “sharpness” at multiple depths of the network. By collecting multiple snapshots throughout the network—especially at earlier layers where features are more generic—IRS helps prevent the network from converging to narrow basins in the loss landscape. Consequently, the learned weights are less vulnerable to internal perturbations, facilitating better generalization and higher inference accuracy under device variations.

Algorithm~\ref{IRS algo} outlines how IRS injects noise $\Delta\mathbf{W} \sim \mathcal{N}(0,\sigma^2)$ not only into the backbone weights but also into the snapshot classifiers, ensuring that all weights experience noise. We then collect the snapshot outputs $Out_n$ in parallel with the backbone’s output $Output$. In this work, we set $\gamma_n=10^{1-n}$, giving shallower representations (i.e., smaller $n$) stronger feedback power. This choice reflects the fact that feature maps closer to the input tend to carry more general information and thus provide more substantial constraints.

Overall, IRS introduces intermediate representation snapshots to serve as internal measurement points, enabling the network to ``see'' how noise affects feature extraction at various depths. This multi-level negative feedback loop promotes flatter, more robust weight and aids in stabilizing the network against NVM device variations. As a result, IRS satisfies the criteria in Section~\ref{sec: NeFT method} and significantly improves inference accuracy for NVCIM accelerators under device variation.

\begin{algorithm}
    \renewcommand{\algorithmicrequire}{\textbf{Input:}}
	\renewcommand{\algorithmicensure}{\textbf{Output:}}
    \caption{IRS~($\mathcal{M}$, $\mathcal{C}$, $\mathbf{W}$, $ep$, $\mathbf{D}$, $\alpha$, $\beta$, $N$)}
    \label{IRS algo}
    \begin{algorithmic}[1]
    \REQUIRE DNN backbone topology $\mathcal{M}$, Classifiers topology $\mathcal{C}$, weight $\mathbf{w}$, \# of training epochs $ep$, dataset $\mathbf{D}$, learning rate $\alpha$, negative feedback coefficient $\beta$, \# of negative feedback $N$;
    \FOR{($i=0$; $i < ep$; $i++$)}
        \FOR{$x$, $\hat{y}$ in $\mathbf{D}$}
            \STATE Sample $\Delta\mathbf{W}_i \sim \mathcal{N}(0,\sigma^2)$;
            \STATE Collect $Output$ and $Out_n$ synchronously with 
            \STATE $\mathcal{M}$($\mathbf{W}$+$\Delta\mathbf{W}_i$,$x$) and $\mathcal{C}$($\mathbf{W'}$+$\Delta\mathbf{W'}_i$,$x$);
            \STATE $\mathcal{O}_\text{feedback}=$ $\sum_{n=1}^{N}$ $10^{1-n}$ $Out_n$;
            \STATE $\mathcal{O}_\text{total} \leftarrow a_{b}\,\mathcal{O}_\text{backbone} - a_{f}\,\beta\,\mathcal{O}_\text{feedback}$;
            \STATE $loss(x,\hat{y}) \leftarrow \text{CrossEntropy}(\mathcal{O}_\text{total}, \hat{y})$;
            \STATE Update $\mathbf{W}$ by gradient descent on $loss$, using the variation-free weight reference.
        \ENDFOR
    \ENDFOR
    \ENSURE Trained, robust DNN backbone $\mathcal{M}(\mathbf{W})$.
    \end{algorithmic}
\end{algorithm}
\vspace{-0.2in}

\subsection{Theoretical Analysis}
\label{sec: theory}

In this section, we provide a interpretation for the convergence behavior of our NeFT-based training algorithms from a control-theoretic perspective. While deep neural networks are generally non-convex, we follow a common simplification~\cite{goodfellow2016deep, dinh2017sharp} used in local stability analysis by assuming that the loss surface is \emph{locally convex} in the vicinity of a relatively good solution.
Let $\mathcal{M}$ be a DNN backbone with weight parameters 
$\mathbf{W} \in \mathbb{R}^{p \times d}$, and let $\Delta \in \mathbb{R}^{p \times d}$ 
be a perturbation matrix. Suppose $\{\mathbf{W}_t\}$ is the sequence of weight matrices 
generated by the proposed algorithm, and let $\alpha$ denote the learning rate. 
The negative feedback training procedure can be described as the following controlled system 
\cite{talukder2023robust}:
\begin{align}\label{eq:control_system}
    \mathbf{W}_{t+1} &= \mathbf{W}_{t} + \alpha F(\mathbf{W}_{t}, \Delta_t),\\
    F(\mathbf{W}_{t}, \Delta_t) &= - \nabla \mathcal{L}_\text{NeFT}(\mathbf{W}_{t}, \Delta_t; \mathbf{x}, \hat{\mathbf{y}}),
\end{align}
where $F$ is the negative gradient of the loss function $\mathcal{L}_\text{NeFT}$, which is guided by the negative feedback principle, concerning the parameters $\mathbf{W}_{t}$ and noise $\Delta_t$. We assume that $\mathcal{L}_\text{NeFT}$ is locally convex and continuously differentiable near the fixed point, and that the eigenvalues of its Hessian $\nabla^2 \mathcal{L}_\text{NeFT}$ are positive and bounded by $m$. This enables us to construct a local contraction mapping for analysis, similar to prior control-inspired training study~\cite{talukder2023robust}. One immediate consequence 
of these assumptions is that $F(\mathbf{W}_{t}, \Delta_t)$ is continuously differentiable. 
We use the contraction mapping principle to show that $\{\mathbf{W}_t\}$ converges to a fixed point.

\smallskip
\noindent
\textbf{Fixed Point Definition.}
Let $\mathbf{W}^*$ be the fixed point to which $\{\mathbf{W}_{t}\}$ converges. Define 
$\mathbf{e}_{n} = \mathbf{W}_{n} - \mathbf{W}^*$. $\mathbf{e}_{n}$ is Lipschitz-bounded~\cite{talukder2023robust}. At this fixed point, we have 
$\mathbb{E}_{\Delta}\bigl[F(\mathbf{W}^*, \Delta_t)\bigr] = 0$, which implies
\begin{equation}
    \mathbb{E}_{\Delta}\bigl[\mathbf{W}^*\bigr]
    = \mathbb{E}_{\Delta}\bigl[\mathbf{W}^*\bigr]
    + \alpha \, \mathbb{E}_{\Delta}\bigl[F(\mathbf{W}^*, \Delta_t)\bigr].
\end{equation}

\noindent
Since $\Delta_t$ is always added to $\mathbf{W}_{t}$ and 
$\mathbb{E}[\Delta_t] = 0$, 
we omit $\Delta_t$ in $F(\mathbf{W}^*, \Delta_t)$ for simplicity in subsequent steps.

\smallskip
\noindent
\textbf{Taylor Expansion.}
We expand $F(\mathbf{W}_t)$ around the point $\mathbf{W}^*$:
\begin{equation}\label{eq:taylor}
    F(\mathbf{W}_t) = F'(\mathbf{W}^*)(\mathbf{W}_t - \mathbf{W}^*) + \mathcal{O}\!\bigl(\|\mathbf{W}_t - \mathbf{W}^*\|^2\bigr),
\end{equation}
where $\|\cdot\|$ denotes the operator norm and $F(\mathbf{W}^*) = 0$ at the fixed point. 
Substituting \eqref{eq:taylor} into \eqref{eq:control_system} yields
\begin{equation}
\begin{aligned}
    \mathbf{W}_{t+1} 
    &= \mathbf{W}_{t} + \alpha \, F(\mathbf{W}_t)\\
    &= \mathbf{W}^* + \mathbf{e}_n 
       + \alpha\,F'(\mathbf{W}^*)\bigl(\mathbf{W}_t - \mathbf{W}^*\bigr) \\
    &\quad + \alpha\,\mathcal{O}\!\bigl(\|\mathbf{W}_t - \mathbf{W}^*\|^2\bigr) \\
    &= \mathbf{W}^* 
       + \bigl(I + \alpha\,F'(\mathbf{W}^*)\bigr)\mathbf{e}_n\\ 
    &\quad + \alpha\,\mathcal{O}\!\bigl(\|\mathbf{W}_t - \mathbf{W}^*\|^2\bigr).
\end{aligned}
\end{equation}
Hence, we have $\mathbf{e}_{n+1}$
\begin{equation}
\begin{aligned}
    \mathbf{e}_{n+1} 
    &= \mathbf{W}_{t+1} - \mathbf{W}^*\\
    &= \bigl(I + \alpha\,F'(\mathbf{W}^*)\bigr)\mathbf{e}_n\;+\;\alpha\,\mathcal{O}\!\bigl(\|\mathbf{W}_t - \mathbf{W}^*\|^2\bigr).
\end{aligned}
\end{equation}

\noindent
When $\mathbf{e}_n$ is sufficiently small, 
$\alpha\,\mathcal{O}\!\bigl(\|\mathbf{W}_t - \mathbf{W}^*\|^2\bigr)$ becomes negligible, 
and the term $\bigl(I + \alpha\,F'(\mathbf{W}^*)\bigr)$ dominates. We next show that 
$\bigl(I + \alpha\,F'(\mathbf{W}^*)\bigr)$ forms a contraction, ensuring $\mathbf{e}_{n+1} \to 0$.

\smallskip
\noindent
\textbf{Contraction Mapping.}
Note that $F(\mathbf{W}_t) = -\nabla \mathcal{L}_\text{NeFT}(\mathbf{W}_t)$, 
so $F'(\mathbf{W}_t) = -\nabla^2 \mathcal{L}_\text{NeFT}(\mathbf{W}_t)$. 
It suffices to show $\|I + \alpha\,F'(\mathbf{W}^*)\| < 1$, 
or $\|I - \alpha\,\nabla^2 \mathcal{L}_\text{NeFT}(\mathbf{W}^*)\| < 1$. 
Given that the Hessian $\nabla^2 \mathcal{L}_\text{NeFT}(\mathbf{W}^*)$ has positive eigenvalues 
bounded by $m$, we choose $0 < \alpha < \tfrac{2}{m}$. This ensures 
\vspace{-0.1in}
\begin{equation}
    \|I - \alpha\,\nabla^2 \mathcal{L}_\text{NeFT}(\mathbf{W}^*)\| < 1,
\end{equation}
making $\bigl(I - \alpha\,\nabla^2 \mathcal{L}_\text{NeFT}\bigr)$ a strict contraction. 
\noindent
As a result, 
\vspace{-0.1in}
\begin{equation}
    \lim_{n \to \infty} \mathbf{e}_{n+1} = 0,
\end{equation}
indicating that the iteration converges to the fixed point $\mathbf{W}^*$. Unlike traditional gradient descent methods, NeFT introduces feedback signals from intermediate results (e.g., variational forwards or intermediate representations) into the training loss. These feedback components constrain the update direction based on the network’s internal state, forming a closed-loop adjustment mechanism. This feedback effectively improves the contraction behavior near stable points by suppressing gradient divergence under device variation. By introducing negative feedback signals into $\mathcal{L}_{\text{NeFT}}$ (where $F(\Delta_t)\neq 0$), we effectively reduce $\|I - \alpha\,\nabla^2 \mathcal{L}_{\text{NeFT}}(\mathbf{W}^*)\|$ compared to the traditional method. This leads to a faster and more stable convergence process. Therefore, NeFT enables a form of \emph{self-regulated learning} in noisy environments by dynamically adjusting the optimization path through internal feedback. The model converges not solely due to external gradients, but also due to its own feedback-constrained structure, which enhances robustness and stability under device variation.
\vspace{-0.2in}

\section{Experiments}\label{sec:experiments}

The primary consequence of device variation is that the conductance of the NVM device deviates from its intended (or target) value. Device-to-device and cycle-to-cycle variations emerge during the programming process, resulting in stochastic disturbances in conductance. Consequently, the weight values stored on the accelerators are not ideal but become statistically distributed, ultimately degrading the inference accuracy. In this section, we first discuss our noise model, which clarifies the relationship between device variation and the noise introduced into the weights during training. After formally defining our problem, we compare our method with the state-of-the-art (SOTA) baseline across various models and datasets. Our results demonstrate improved inference accuracy on the NVCIM DNN accelerator platform. Furthermore, our method improves prediction confidence and increases the likelihood of model convergence under device variations.
\vspace{-0.2in}

\subsection{Modeling Weight Perturbation}\label{sec:noise model}

Without loss of generality, we focus on variations arising from the programming process of NVM devices, where the programmed conductance deviates from its intended (target) value, but our model can be used in other variations. In what follows, we establish a mathematical formulation for the mapping of DNN weights to NVM device states and then explicitly model the effect of these variations on weight values.

Let $\mathcal{W} \subset \mathbb{R}$ represent the set of floating-point DNN weights, and let $\max|\mathcal{W}|$ denote the maximum absolute weight. For a quantized network with $M$-bit precision, each weight $w \in \mathcal{W}$ is mapped to one of $2^M$ uniform levels. We define the \emph{desired} (or \emph{target}) quantized weight value $\bar{\mathcal{W}_d}$ as
\begin{equation}\label{eq:desiredWeight}
    \bar{\mathcal{W}_d} \;=\; \frac{\max|\mathcal{W}|}{2^M - 1} \,
    \sum_{i=0}^{M-1} m_{i}\,\times 2^{\,i},
\end{equation}
where $m_i \in \{0,1\}$ is the $i$-th bit of the quantized representation, and $\tfrac{\max|\mathcal{W}|}{2^M - 1}$ rescales the bit pattern into the correct real range. Negative weights can be handled by mapping them to a separate crossbar array (or equivalently processing the sign bit separately).

Suppose each NVM device can store $K$ bits of conductance (i.e., $2^K - 1$ discrete conductance levels). A single $M$-bit weight is thus mapped onto $M/K$ such devices\footnote{For simplicity, we assume $M$ is a multiple of $K$.}. Let $\bar{g}_j$ be the target conductance level of the $j$-th device, determined by its local $K$-bit segment:
\vspace{-0.1in}
\begin{equation}\label{eq:deviceConductance}
    \bar{g}_j \;=\; \sum_{i=0}^{K-1} m_{jK + i}\,\times 2^{\,i}.
\end{equation}
Ideally, each device’s programmed conductance matches its intended value $\bar{g}_j$ exactly. In practice, the actual conductance $g_j$ deviates from $\bar{g}_j$. We model this deviation $\Delta g$ as a Gaussian random variable, $\Delta g \sim \mathcal{N}\!\bigl(0,\sigma_d^2\bigr)$, where $\sigma_d$ represents the standard deviation. Formally,
\vspace{-0.1in}
\begin{equation}\label{eq:actualConductance}
    g_j \;=\; \bar{g}_j \;+\;\Delta g_j, 
    \quad \Delta g_j \sim \mathcal{N}(0,\sigma_d^2).
\end{equation}
The parameter $\sigma_d \leq 0.4$ reflects a practical range in which device-level optimizations (e.g., selective write-verify~\cite{shim2020two}) can hold the variation in check. In typical scenarios, the device programming variation $\sigma_d$ is reported to fall below 0.2~\cite{wang2022device, doevenspeck2021oxrram, yan2022swim}. However, to account for additional non-idealities such as aging, retention loss, and drift, we adopt a more conservative upper bound of 0.4 as used in prior works~\cite{huang2022efficient, yu2023cola, eldebiky2023correctnet}.
Each $M$-bit weight is split across $M/K$ devices. Let $\Delta g_j$ be the random deviation in the $j$-th device’s conductance. The actual programmed weight $\mathcal{W}_p$ that emerges on the accelerator can be expressed as
\vspace{-0.2in}
\begin{equation}
    \mathcal{W}_p \;=\; \bar{\mathcal{W}_d} \;+\;\underbrace{\frac{\max|\mathcal{W}|}{2^M - 1}}_{\text{quant. scaling}} 
      \sum_{j=0}^{\tfrac{M}{K}-1} \!\Delta g_j\,\times2^{jK}
\end{equation}

Following the setup in \cite{shim2020two, jiang2020device}, we set $K=2$. The total $M$ bits used per weight depends on the DNN configuration; in this work, we choose $M=8$ bits per weight (thus each weight is mapped onto 4 devices). We maintain the relative standard deviation of no more than 0.2, a range deemed feasible through device-level improvements such as write-verify. Overall, these settings match previous works, forming the basis for our subsequent experiments.
\vspace{-0.2in}

\subsection{Experimental Setup}
\label{sec:exp setup}

We conduct all experiments in the PyTorch framework, utilizing NVIDIA XPs and P100s for accelerated computation. To thoroughly assess our NeFT method, we benchmark it against three primary baselines~\cite{jiang2020device, yang2022tolerating, eldebiky2023correctnet}: 
\emph{Vanilla training (W/O Noise)}, wherein the model is trained under optimal hyperparameters without any injected noise;
\emph{Gaussian noise-injection training (W/ Noise)}, which introduces Gaussian noise correlated to the target device variation during training;
\emph{CorrectNet}, in which the objective is augmented by a modified Lipschitz constant regularization term to enhance robustness.

We did not compare with orthogonal or hardware-focused solutions such as TSB or selective write-verify, given that our approach is complementary and can be integrated with them if desired. An exhaustive search reveals that a negative feedback coefficient $\beta$ drawn from $\{10^{-1}, 10^{-2}, 10^{-3}, 10^{-4}\}$ reliably achieves near-optimal performance. The choice of the decay factor $\gamma$ is based on a heuristic approach. We observed that applying cascaded feedback signals with geometrically decaying factors (e.g., $10^{-1}, 10^{-2}$) helps stabilize training. This setup preserves dominant contributions from recent representations while retaining weaker historical influence.

For noise-related hyperparameters, we set \textit{start} = 0 and \textit{end} = $2 \times \sigma_d$, and systematically vary $\sigma_d$ within the range described in Section~\ref{sec:noise model}. We further assign $a_b = a_f = (N + 1)^{-1}$, where $N$ is the number of negative feedback signals introduced by NeFT. All other training hyperparameters, including the learning rate, batch size, and scheduling strategies, adhere to established best practices for noise-free model training, ensuring that our experiments focus on device variation.

Each method tries to train the model to converge on GPUs before mapping its final parameters to the NVCIM accelerator; subsequent inference accuracy is evaluated under injected noise to emulate real-world device variations on the NVCIM accelerator. Unless otherwise noted, all results represent the average of at least five independent runs. Due to the random variability of the generated noise in the test, we employ a Monte Carlo (MC) simulation comprising 5,000 runs, providing a statistical precision that yields a 95\% confidence interval of $\pm 0.01$, under the central limit theorem.
\vspace{-0.2in}

\subsection{Effectiveness on MNIST}

\begin{figure}
  \centering
  \includegraphics[scale=0.5]{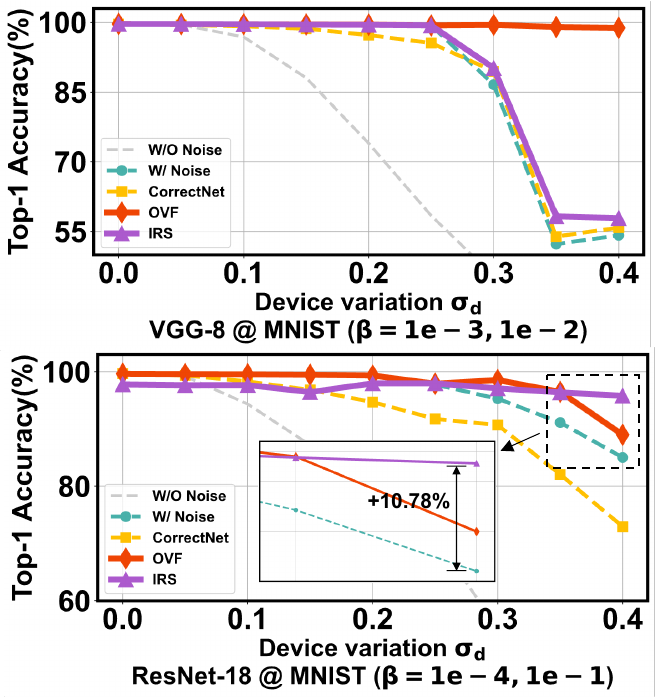}
  \caption{Effectiveness of NeFT, implemented through the proposed OVF and IRS instances. The plots show average inference accuracy under noisy conditions on (a) VGG-8 and (b) ResNet-18 backbones, evaluated on the MNIST dataset across varying $\sigma_d$ values. Dashed lines represent baseline methods; the negative feedback coefficients $\beta$ are for OVF or IRS.}
  \label{fig:mnist_acc}
  \vspace{-0.2in}
\end{figure}

We begin our discussion by demonstrating the effectiveness of the NeFT approach using VGG-8 and ResNet-18 architectures on the MNIST dataset. Under ideal conditions with no device variations, these models achieve respective accuracy rates of 99.64\% and 99.65\%. VGG-8 and ResNet-18 contain $1.29 \times 10^7$ and $1.15 \times 10^7$ weight parameters.

Although the typical relative standard deviation of device conductance is below 0.2 before applying write-verify operations, emerging experimental technologies can exhibit more significant variations. To demonstrate the universal applicability of the NeFT approach, we thus evaluate its two implementations—IRS and OVF—under an extended range of device variations. This broader scenario encompasses real-world factors such as experimental technologies, device aging, and other non-ideal degradation effects, offering a more comprehensive evaluation of method robustness.

Figure~\ref{fig:mnist_acc} compares the top‐1 inference accuracy of five methods (W/O Noise, W/ Noise, CorrectNet, OVF, and IRS) across different device variation levels. All models maintain near‐perfect accuracy for small $\sigma_d$, indicating that minor variations do not significantly degrade performance. For small $\sigma_d$, the improvement from NeFT remains minimal since MNIST is relatively simple for both backbones, and minor weight perturbations stay within the tolerance range of these original networks. However, as $\sigma_d$ increases, the vanilla training approach (W/O Noise) experiences a substantial drop in accuracy, emphasizing its vulnerability to moderate device variations. Gaussian noise‐injection (W/ Noise) and CorrectNet degrade more gently but still encounter notable performance drops when $\sigma_d$ becomes large. By contrast, OVF and IRS exhibit markedly higher robustness. On the VGG-8 backbone, for instance, NeFT(OVF) achieves an absolute accuracy improvement of up to 45.08\% compared to the best baseline, CorrectNet. This improvement arises from two factors. First, NeFT enhances the network’s tolerance to device variations. Second, especially in the OVF setting, NeFT aids in achieving more reliable convergence during training, thereby boosting average accuracy. Meanwhile, on the ResNet-18 backbone, IRS delivers an absolute accuracy gain of up to 10.78\% over the best baseline (W/ Noise), as illustrated in the figure inset.
\vspace{-0.2in}

\subsection{Effectiveness on CIFAR-10}

\begin{figure}[t]
  \centering
  \includegraphics[scale=0.5]{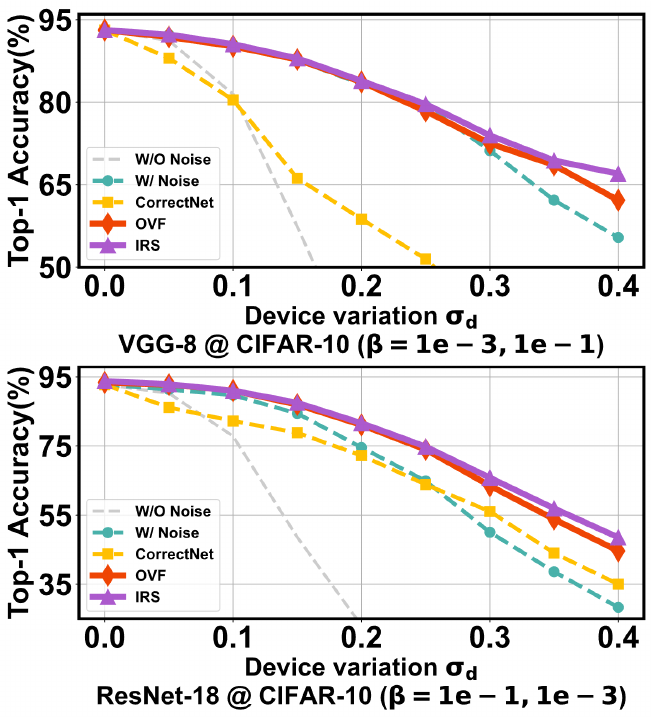}
  \caption{Effectiveness of NeFT, implemented through the proposed OVF and IRS instances. The plots show average inference accuracy under noisy conditions on (a) VGG-8 and (b) ResNet-18 backbones, evaluated on the CIFAR-10 dataset across varying $\sigma_d$ values. Dashed lines represent baseline methods; the negative feedback coefficients $\beta$ are for OVF or IRS.}
  \label{fig:cifar10_acc}
  \vspace{-0.2in}
\end{figure}

To further assess the general applicability of our approach, we conduct experiments on the CIFAR-10 dataset using VGG-8 and ResNet-18 backbones. Under noise-free conditions, these models achieve top-1 accuracy rates of 91.84\% and 92.71\%, respectively. Figure~\ref{fig:cifar10_acc} compares the performance of five methods (W/O Noise, W/ Noise, CorrectNet, OVF, and IRS) as the device variation $\sigma_d$ increases to 0.4.

Similar to the MNIST setting, the W/O Noise baseline deteriorates rapidly once $\sigma_d$ increases. Meanwhile, Gaussian noise-injection (W/ Noise) and CorrectNet are more robust than W/O Noise, yet both still suffer significant accuracy drops when $\sigma_d$ becomes large. In contrast, the proposed OVF and IRS demonstrate notably higher resilience across the variation range. On the VGG-8 backbone, for instance, OVF and IRS consistently outperform W/O Noise and CorrectNet. Although W/ Noise follows a smoother decline curve, NeFT(IRS) surpasses it by up to 11.60\% in absolute accuracy. A more obvious trend appears on the ResNet-18 backbone, where IRS and OVF maintain a 5–7\% advantage over the strongest baseline under moderate variations, and this gap widens as the best practice of NeFT achieves up to 13.54\% improvement.
\vspace{-0.2in}

\subsection{Effectiveness on larger datasets}

\begin{figure}
  \centering
  \includegraphics[scale=0.5]{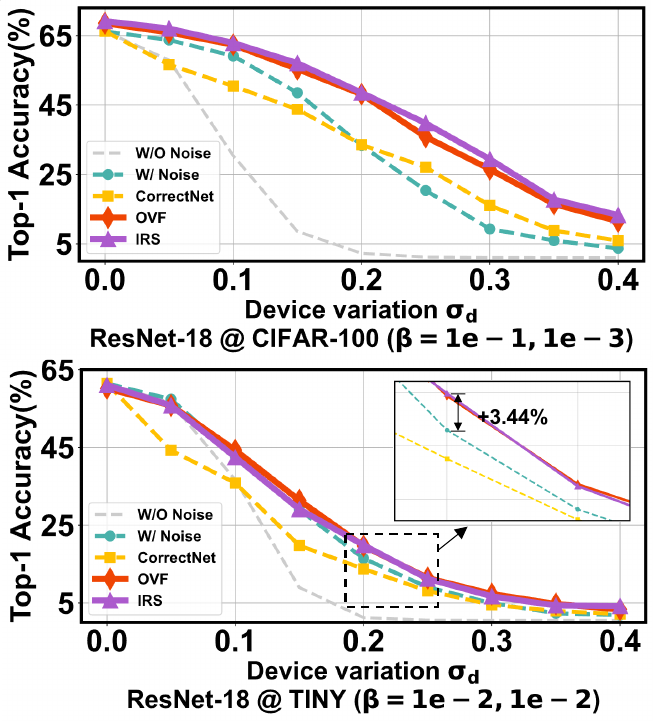}
  \caption{Effectiveness of NeFT, implemented through the proposed OVF and IRS instances. The plots show average inference accuracy under noisy conditions on the ResNet-18 backbone, evaluated on the (a) CIFAR-100 and (b) Tiny ImageNet datasets across varying $\sigma_d$ values. Dashed lines represent baseline methods; the negative feedback coefficients $\beta$ are for OVF or IRS.}
  \label{fig:largedataset_acc}
  \vspace{-0.2in}
\end{figure}

To evaluate our approach on more challenging datasets, we deploy ResNet-18 on both CIFAR-100 and Tiny ImageNet. These datasets offer a larger label space and increased data complexity compared to CIFAR-10 and MNIST. Figure~\ref{fig:largedataset_acc} presents the top-1 accuracy as the device variation $\sigma_d$ increases. Under noise-free conditions, ResNet-18 achieves accuracy rates of 66.25\% on CIFAR-100 dataset and 61.45\% on Tiny ImageNet dataset, respectively.

On CIFAR-100, NeFT maintains a clear advantage over the baselines across the entire range of $\sigma_d$, outperforming them by several percentage points. In particular, with the IRS configuration, NeFT achieves a notable 14.96\% improvement over the strongest baseline (CorrectNet) at $\sigma_d = 0.2$. These results underscore how negative feedback training facilitates more robust weight updates in the presence of significant device variations, 
leading to higher accuracy even for a complex dataset like CIFAR-100 dataset.

Meanwhile, on Tiny ImageNet, NeFT(OVF) delivers up to a 3.44\% accuracy increase over the best baseline under high noise levels, highlighting its capacity to improve network resilience. However, the absolute accuracy gains on Tiny ImageNet are not as pronounced as those on CIFAR-100, presumably because Tiny ImageNet’s increased complexity—in terms of both the number of classes and more diverse images—may exceed the representational capacity of ResNet-18 under substantial device variations. Nonetheless, the negative feedback mechanism still consistently mitigates performance degradation, reinforcing NeFT’s versatility for larger datasets. These findings collectively illustrate the efficacy of our approach in handling device variation, although higher dataset complexity can limit the relative margin of improvement.
\vspace{-0.2in}

\subsection{Description for Devices With Nonuniform Variations}\label{sec:nonuniform models}

In the previous sections, we demonstrated the effectiveness of NeFT under varying device variations; however, all prior noise models assumed a uniform device variation (or a RRAM-based device model \emph{RRAM1}), where the standard deviation $\sigma_d$ remains fixed regardless of device conductance values. To further showcase the robust capability of our NeFT method, we now introduce three additional device models for evaluation. Similar to earlier settings, each device has four possible conductance levels (0, 1, 2, and 3), but these new models allow $\sigma_d$ to vary as the device conductance $g_j$ changes. This setup offers a more realistic representation of hardware behavior and provides deeper insights into NeFT’s ability to handle non-uniform noise distributions.

Specifically, three additional devices can be modeled as:
\vspace{-0.1in}
\begin{equation}
    \sigma_{d} = 
    \begin{cases}
        1, & g_j = 0, 3\\
        t, & g_j = 1, 2
    \end{cases}
\end{equation}
where $t=4$ for the second RRAM-based model. This model is abstracted from empirical data in~\cite{liu2023architecture} and referred as \emph{RRAM2}. For the second and third devices derived from a FeFET model~\cite{wei2022switching}, denoted as 
\emph{FeFET1} and \emph{FeFET2}, we have $t=2$ and $t=6$.
By doing so, we capture a broader range of real-world non-idealities from device variability. Having defined the three additional device models, we now integrate them into our hardware simulation environment to evaluate the robustness of NeFT against non-uniform variations.
\vspace{-0.2in}

\subsection{Effectiveness under Nonuniform Device Variations}
\begin{table*}[]
\centering
\caption{Comparison of accuracy (\%) and device variation ($\sigma_d$) between NeFT (OVF and IRS) and baseline methods (W/ Noise and CorrectNet) on ResNet-18 for the CIFAR-10 dataset. Three nonuniform device models are considered, as specified in Section~\ref{sec:nonuniform models}.}
\label{tab:different devices acc}
\resizebox{\textwidth}{!}{%
\begin{tabular}{lcccccccccc}
\hline
\multicolumn{1}{c}{\multirow{2}{*}{Device}} & \multicolumn{1}{l}{\multirow{2}{*}{Methods}} & \multicolumn{9}{c}{Device Variation ($\sigma_d$)} \\
\multicolumn{1}{c}{} & \multicolumn{1}{l}{} & 0.0 & 0.05 & 0.1 & 0.15 & 0.2 & 0.25 & 0.3 & 0.35 & 0.4 \\ \hline
\multirow{4}{*}{RRAM2} & W/ noise & \multirow{4}{*}{\begin{tabular}[c]{@{}c@{}}$\uparrow$\\ 92.71$\pm$0.02\\ $\downarrow$\end{tabular}} & 92.30$\pm$0.20 & 91.35$\pm$0.43 & 90.53$\pm$0.77 & 87.97$\pm$1.52 & 85.19$\pm$2.20 & 80.78$\pm$3.95 & 76.41$\pm$5.09 & 71.05$\pm$6.47 \\
 & CorrectNet &  & 86.47$\pm$0.97 & 81.45$\pm$2.75 & 76.57$\pm$4.42 & 71.65$\pm$5.66 & 65.01$\pm$7.24 & 60.52$\pm$7.71 & 52.54$\pm$8.77 & 46.26$\pm$8.98 \\
 & NeFT(OVF) &  & 93.14$\pm$0.19 & \textbf{92.45$\pm$0.32} & \textbf{91.14$\pm$0.62} & 89.68$\pm$1.06 & 87.78$\pm$1.58 & 84.18$\pm$3.03 & 81.45$\pm$3.61 & \textbf{78.82$\pm$4.54} \\
 & NeFT(IRS) &  & \textbf{93.23$\pm$0.15} & 92.39$\pm$0.36 & 90.98$\pm$0.80 & \textbf{89.71$\pm$1.10} & \textbf{87.83$\pm$1.70} & \textbf{85.29$\pm$2.48} & \textbf{81.57$\pm$3.88} & 76.88$\pm$5.22 \\ \hline
\multirow{4}{*}{FeFET1} & W/ noise & \multirow{4}{*}{\begin{tabular}[c]{@{}c@{}}$\uparrow$\\ 92.71$\pm$0.02\\ $\downarrow$\end{tabular}} & 92.15$\pm$0.21 & 90.75$\pm$0.63 & 88.47$\pm$1.15 & 82.58$\pm$2.74 & 77.54$\pm$4.63 & 68.51$\pm$6.90 & 59.09$\pm$8.87 & 51.14$\pm$9.78 \\
 & CorrectNet &  & 86.35$\pm$0.90 & 81.86$\pm$2.62 & 77.91$\pm$3.68 & 72.71$\pm$5.19 & 64.38$\pm$7.80 & 59.77$\pm$8.22 & 50.51$\pm$9.50 & 48.36$\pm$8.83 \\
 & NeFT(OVF) &  & 93.06$\pm$0.20 & 91.48$\pm$0.47 & 89.97$\pm$0.91 & 86.91$\pm$1.98 & \textbf{83.23$\pm$3.33} & \textbf{78.50$\pm$4.76} & 68.42$\pm$7.59 & \textbf{65.69$\pm$8.74} \\
 & NeFT(IRS) &  & \textbf{93.15$\pm$0.17} & \textbf{91.86$\pm$0.43} & \textbf{89.99$\pm$0.97} & \textbf{87.17$\pm$1.65} & 82.96$\pm$2.90 & 76.07$\pm$5.40 & \textbf{71.25$\pm$6.75} & 63.34$\pm$8.73 \\ \hline
\multirow{4}{*}{FeFET2} & W/ noise & \multirow{4}{*}{\begin{tabular}[c]{@{}c@{}}$\uparrow$\\ 92.71$\pm$0.02\\ $\downarrow$\end{tabular}} & 92.17$\pm$0.18 & 91.52$\pm$0.47 & 90.32$\pm$0.80 & 88.91$\pm$1.28 & 86.76$\pm$1.82 & 83.77$\pm$3.06 & 80.43$\pm$3.66 & 76.51$\pm$5.41 \\
 & CorrectNet &  & 85.83$\pm$1.04 & 81.78$\pm$2.78 & 77.21$\pm$3.93 & 70.85$\pm$5.97 & 64.14$\pm$8.03 & 60.70$\pm$8.29 & 52.67$\pm$8.59 & 43.48$\pm$9.14 \\
 & NeFT(OVF) &  & 92.88$\pm$0.16 & 92.45$\pm$0.32 & 91.46$\pm$0.56 & \textbf{90.32$\pm$0.89} & 88.26$\pm$1.63 & 86.37$\pm$2.13 & 82.58$\pm$3.74 & 79.99$\pm$4.57 \\
 & NeFT(IRS) &  & \textbf{93.41$\pm$0.13} & \textbf{92.64$\pm$0.30} & \textbf{91.53$\pm$0.58} & 90.28$\pm$0.96 & \textbf{88.56$\pm$1.59} & \textbf{86.81$\pm$2.01} & \textbf{83.41$\pm$3.36} & \textbf{80.03$\pm$4.59} \\ \hline
\end{tabular}%
}
\vspace{-0.2in}
\end{table*}

Table~\ref{tab:different devices acc} presents a detailed comparison of average inference accuracy (\%) and device variation ($\sigma_d$) for ResNet-18 on the CIFAR-10 dataset, evaluated under three nonuniform device models described in Section~\ref{sec:nonuniform models}. The accuracy values are reported in mean$\pm$standard-deviation form over 5,000 Monte Carlo simulations and 5 separate runs. A value of $\sigma_d = 0.0$ corresponds to an idealized scenario with no weight noise during training and inference. 

Notably, under all three device models (RRAM2, FeFET1, and FeFET2), our proposed NeFT instances (OVF and IRS) consistently exhibit higher robustness compared to the baseline methods (W/ noise and CorrectNet). For instance, in the RRAM2 device setting, NeFT sustains accuracy above $80\%$ even at $\sigma_d = 0.35$, whereas the strongest baseline drops to $76.41\%$. A similar trend is observed in the FeFET1 model, where NeFT(IRS) outperforms CorrectNet by approximately $10\%$ at high noise levels. In the FeFET2 device setting, NeFT(IRS) maintains accuracy around $86\%$ at $\sigma_d = 0.3$, exceeding the best baseline by a clear margin. These results confirm that NeFT effectively adapts to varying degrees of non-uniform conductance noise, highlighting its advantage in mitigating performance degradation even under significant device variations in NVCIM accelerators.
\vspace{-0.2in}

\subsection{Generalization under Different Variation}
\label{sec:log-normal noise}

Based on the device researches~\cite{yan2022computing, yan2023improving, qin2020design, qin2024tsb, doevenspeck2021oxrram}, the device variation shows a Gaussian distribution. But for some cases like HRS distribution in NVM devices and reading drift, the device variation can also be an asymmetric distribution~\cite{qin2020design, huang2022efficient}. To further evaluate the generalizability of NeFT beyond standard Gaussian variation, we conduct an additional experiment using a non-Gaussian, asymmetric exponential noise model. This model, commonly adopted in NVCIM accelerator studies~\cite{yu2023cola, huang2022efficient}, reflects the asymmetric and temporally correlated characteristics observed in NVM devices.
Following the modeling approach in~\cite{yu2023cola}, we simulate state-dependent perturbations by applying log-normal multiplicative noise to the weights. Formally, the perturbed weights $\mathcal{W}_p$ are computed as $\mathcal{W}_p = \mathcal{W} \odot e^{v}$ where $\odot$ denotes element-wise multiplication and $v \sim \mathcal{N}(0, \gamma^2)$ is an i.i.d. Gaussian matrix of the same shape as $\mathcal{W}$. This formulation produces a right-skewed distribution over the weights, where most values remain close to their original magnitude, but occasional large deviations (long-tail behavior) can occur—consistent with drift effects reported in emerging NVM technologies~\cite{qin2020design, antolini2023combined}.
We evaluate NeFT (both OVF and IRS) under this noise model using the ResNet-18 architecture on the CIFAR-10 dataset. The noise scale $\sigma_d$ is aligned with the standard deviation values used in previous Gaussian experiments to ensure fair comparison. The results are summarized in Table~\ref{tab:different variation}, showing that while noise-injection training and CorrectNet suffer noticeable degradation under log-normal noise, NeFT (OVF and IRS) consistently maintains high robustness (exceeding 90\% accuracy) with only minor accuracy drops. 
These findings confirm that NeFT generalizes well to asymmetric, multiplicative perturbations, and not merely to symmetric additive noise, reinforcing its applicability to practical NVM-based CIM systems under complex device variation conditions.

\subsection{Uncertainty and Convergence}

\begin{figure}[]
  \centering
  \includegraphics[scale=0.31]{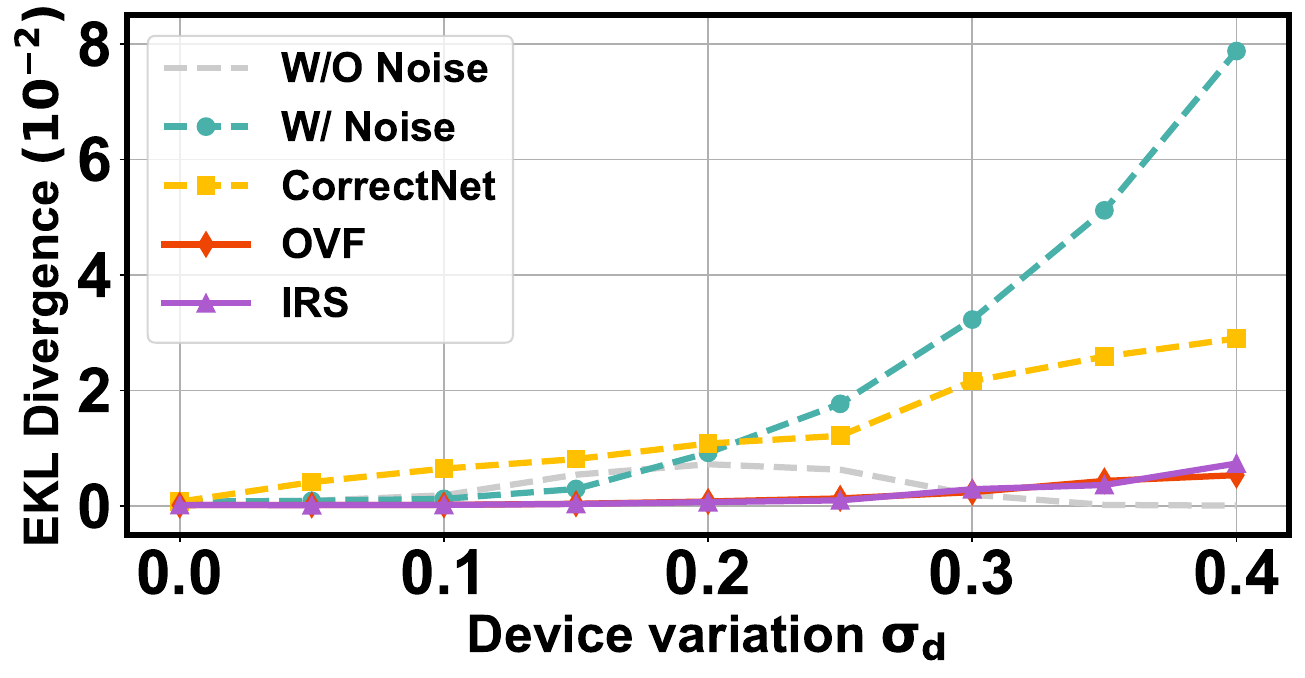}
  \caption{Average expected Kullback–Leibler (EKL) divergence of NeFT (OVF and IRS) and baseline methods (W/O Noise, W/ Noise, CorrectNet) on ResNet-18 for the CIFAR-10 dataset, evaluated under varying device variation $\sigma_d$. To ensure a fair comparison and isolate the effect of accuracy, the EKL divergence is averaged only over correct predictions. Lower EKL values correspond to higher prediction confidence and thus indicate better performance.}
  \label{fig:EKL}
  \vspace{-0.3in}
\end{figure}

Device variations can significantly increase epistemic uncertainty, manifesting as higher output uncertainty in deep neural network predictions. To formally quantify this phenomenon, we employ the Expected Kullback–Leibler (EKL) Divergence~\cite{gawlikowski2023survey}, which measures the discrepancy between a model’s predicted probability distribution and the true label distribution (a one-hot vector). A lower EKL divergence indicates greater prediction confidence. Crucially, we compute the Kullback–Leibler Divergence \emph{only} for correctly classified samples during noisy inference, thereby disentangling accuracy from uncertainty. This approach ensures that a model’s confidence level is not conflated with its failure to predict the correct label. Figure~\ref{fig:EKL} illustrates the averaged EKL divergence over each correct prediction sample, comparing our NeFT approach (OVF and IRS) against three baselines: W/O Noise, W/ Noise, and CorrectNet. While W/ Noise and CorrectNet improve accuracy relative to the W/O Noise baseline, they also elevate model uncertainty by a noticeable margin. In contrast, NeFT (OVF and IRS) not only sustains higher accuracy than baselines but also maintains lower output uncertainty. Notably, when device variation becomes large—beyond the regime where vanilla training can produce meaningful predictions—the EKL divergence for the vanilla training (W/O Noise) baseline may appear slightly lower than that of NeFT, simply because the baseline’s predictions are closer to random guesses and hence devoid of informative meaning.

Severe device variation, as can arise in real-world scenarios like manufacturing inconsistencies, device aging, or environmental fluctuations, also poses challenges for model convergence. Our results indicate that the reduced uncertainty from NeFT assists in stabilizing training and curbing convergence failures. For instance, consider the case of training VGG-8 on MNIST at $\sigma_d = 0.35$. Across 10 independent runs, W/ noise baseline exhibits 6 non-convergent models,\footnote{We define a model as \emph{non-convergent} if its accuracy lags by more than 10\% relative to the mean accuracy observed across multiple runs.} whereas NeFT(OVF) shows no such failures and NeFT(IRS) exhibits only 2. Such stability translates into higher final accuracies, helping to explain the pronounced gains achieved by NeFT(OVF) for VGG-8 in MNIST, as illustrated in Figure~\ref{fig:mnist_acc}. Beyond reinforcing accuracy, these findings highlight the broader value of controlling epistemic uncertainty to ensure robust and consistent performance, even under substantial deviations in device conductance.
\vspace{-0.2in}
\begin{table*}[]
\centering
\caption{Comparison of accuracy (\%) and device variation ($\sigma_d$) between NeFT (OVF and IRS) and baseline methods (W/ Noise and CorrectNet) on ResNet-18 for the CIFAR-10 dataset. The log-normal device variation is considered, as specified in Section~\ref{sec:log-normal noise}.}
\label{tab:different variation}
\resizebox{\textwidth}{!}{%
\begin{tabular}{cccccccccc}
\hline
\multirow{2}{*}{Methods} & \multicolumn{9}{c}{Device Variation ($\sigma_d$)} \\
 & 0.0 & 0.05 & 0.1 & 0.15 & 0.2 & 0.25 & 0.3 & 0.35 & 0.4 \\ \hline
W/ noise & \multirow{4}{*}{\begin{tabular}[c]{@{}c@{}}$\uparrow$\\ 92.71$\pm$0.02\\ $\downarrow$\end{tabular}} & 92.68$\pm$0.08 & 92.86$\pm$0.18 & 92.54$\pm$0.22 & 92.05$\pm$0.31 & 91.89$\pm$0.35 & 91.10$\pm$0.57 & 89.63$\pm$0.69 & 88.24$\pm$0.85 \\
CorrectNet &  & 87.79$\pm$0.47 & 84.73$\pm$1.51 & 81.74$\pm$2.64 & 78.86$\pm$3.86 & 75.10$\pm$4.38 & 70.59$\pm$6.29 & 66.10$\pm$7.36 & 62.13$\pm$8.46 \\
NeFT(OVF) &  & \textbf{93.69$\pm$0.07} & \textbf{93.39$\pm$0.12} & \textbf{93.43$\pm$0.18} & 92.84$\pm$0.29 & \textbf{92.70$\pm$0.29} & 91.44$\pm$0.39 & 91.57$\pm$0.98 & 90.83$\pm$0.86 \\
NeFT(IRS) &  & 93.49$\pm$0.09 & 93.19$\pm$0.11 & 93.30$\pm$0.18 & \textbf{93.15$\pm$0.17} & 92.50$\pm$0.26 & \textbf{92.37$\pm$0.43} & \textbf{91.78$\pm$0.51} & \textbf{91.46$\pm$0.62} \\ \hline
\end{tabular}%
}
\vspace{-0.2in}
\end{table*}

\subsection{Ablation study}

In this section, we present a series of ablation studies that shed further light on the behavior and design choices of NeFT. Unless otherwise stated, all experiments employ ResNet-18 on the CIFAR-10 dataset with a device variation level of $\sigma_d = 0.3$. Our goal is to verify how different configurations within the NeFT framework affect robustness against device variation.

\begin{figure}[]
  \centering
  \includegraphics[scale=0.32]{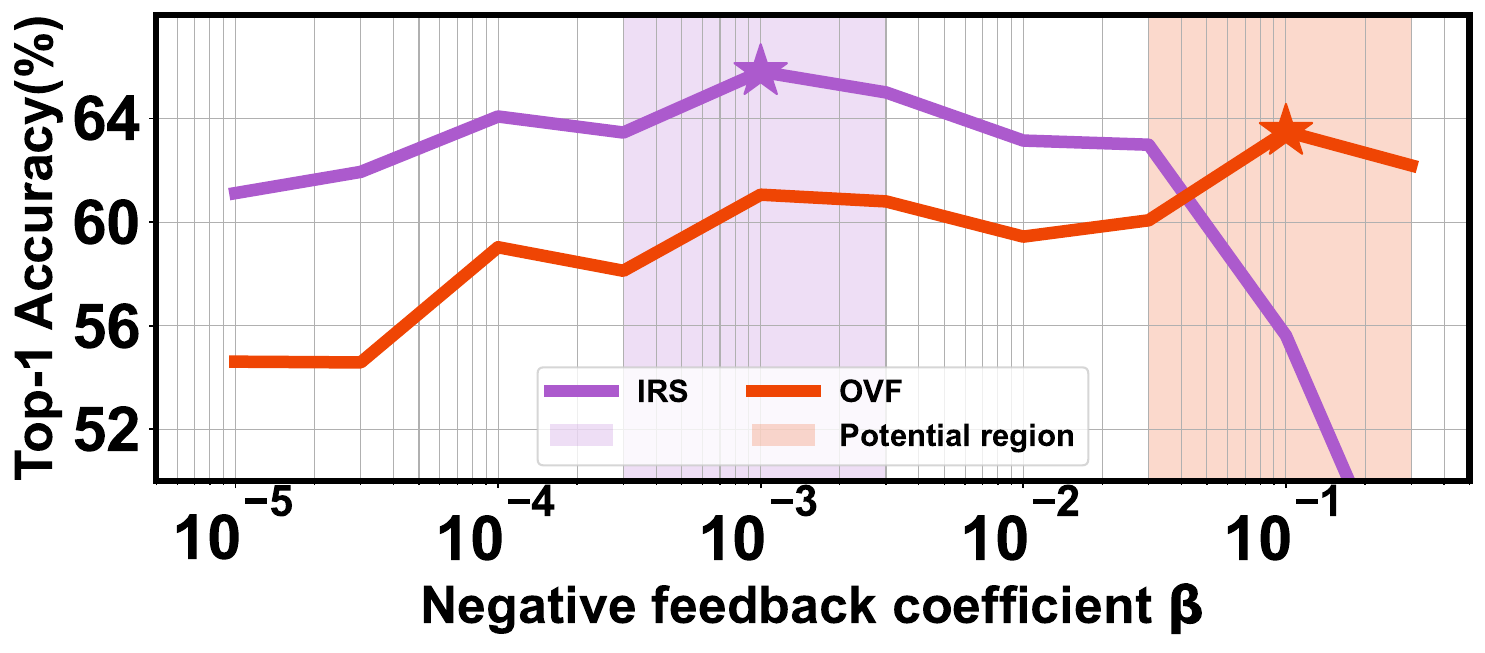}
  \caption{Noisy inference accuracy under different values of the negative feedback coefficient $\beta$, evaluated on ResNet-18 with the CIFAR-10 dataset and device variation level $\sigma_d = 0.3$. The shaded region indicates a stable operating range for $\beta$, where NeFT maintains high robustness. Results demonstrate that $\beta = 10^{-1}$ and $10^{-3}$ for OVF and IRS offer a good balance between stability and effectiveness.
}
  \label{fig:beta_param}
  \vspace{-0.2in}
\end{figure}

\textbf{Negative feedback coefficient $\beta$.}
As stated in Section~\ref{sec:exp setup}, a four-step grid search over $\beta$ is recommended for each method, backbone, and dataset. To evaluate the sensitivity of NeFT to the negative feedback coefficient, we conduct a parameter sweep experiment using ResNet-18 on CIFAR-10 under a fixed device variation level of $\sigma_d = 0.3$. 
As shown in Figure~\ref{fig:beta_param}, the noisy inference accuracy remains relatively stable within the shaded regions, where $\beta$ values yield robust and consistent performance. This indicates that NeFT is not overly sensitive to the precise value of $\beta$ within this operating range. To balance generality and ease of deployment, we select $\beta = 10^{-1}$ for OVF and $\beta = 10^{-3}$ for IRS on ResNet-18 with CIFAR-10.
When $\beta$ is too small (e.g., $10^{-5}$), the feedback signal becomes negligible, reducing NeFT to an unregularized form. Conversely, when $\beta$ is too large (e.g., $>10^{-1}$), training becomes unstable due to excessive gradient modulation. These results confirm the existence of a robust, practical range for setting $\beta$. 

\textbf{Effect of Varying the Number of Negative Feedback Outputs.}
A key hyperparameter in NeFT is the number of negative feedback outputs, $N$, which determines the number of distinct feedback signals (snapshots or oriented forwards) that contribute to the final feedback. Table~\ref{tab:number of N} summarizes the results for OVF and IRS, illustrating how increasing $N$ influences final inference accuracy under noise. When $N$ increases from 1 to 3 (for OVF) or 4 (for IRS), we observe consistent improvements in accuracy, indicating that additional feedback signals enhance the model's ability to counteract device variations. Although a further increase in $N$ can yield a marginal accuracy boost (1–2\% on average), it comes at the cost of additional training overhead, including longer run times and higher computational resource usage. In practical scenarios, we therefore settle on $N=3$ for OVF and $N=4$ for IRS (both marked with $^*$ in Table~\ref{tab:number of N}) as a balance between performance gains and efficiency.
\begin{table}[]
\centering
\caption{Impact of the number of negative feedback signals $Out_n$ on ResNet-18 with CIFAR-10 ($\sigma_d=0.3$). The setting marked with $^*$ achieve the best trade-off between accuracy and computational overhead.}
\label{tab:number of N}
\resizebox{\columnwidth}{!}{%
\begin{tabular}{cccccc}
\hline
\multirow{2}{*}{Methods} & \multicolumn{5}{c}{Number of feedback signals} \\
 & 1 & 2 & 3 & 4 & 5 \\ \hline
NeFT(OVF) & 57.24 & 61.60 & 63.50$^*$ & 65.67 & - \\
NeFT(IRS) & 55.93 & 61.58 & 65.43 & 65.81$^*$ & 66.44 \\ \hline
\end{tabular}%
}
\vspace{-0.2in}
\end{table}

\textbf{Reverse Decay Factors.}
In this experiment, we reverse the direction of the decay factors $\gamma_n$, so that feedback components exhibiting larger deviations from the target receive smaller $\gamma_n$ values. This arrangement stands in contrast to the original NeFT configuration, which assigns stronger negative feedback to higher‐deviation signals. As shown in our results, reversing these factors causes OVF and IRS accuracies to drop to 46.01\% and 44.90\%, respectively, from 63.50\% and 65.81\% under the standard NeFT setup. Such a substantial reduction validates our original design, underscoring the importance of allocating greater influence 
to signals reflecting larger deviations. Intuitively, downplaying high‐deviation feedback diminishes the model’s corrective response to extreme errors or noise, making it more likely to converge to suboptimal weights. By contrast, when larger deviations carry a proportionally stronger impact, the network is pushed harder to remedy severe perturbations, resulting in a more robust and stable solution under device variations.

\textbf{Reverse Oriented $\sigma_d$.}
This experiment focuses on OVF, examining what happens when the oriented variational forwards generate higher‐quality outputs rather than the noisier ones originally intended to steer the backbone’s optimization. Concretely, we reverse the sign of $\Delta\sigma_d$ from $+0.05$ to $-0.05$, causing the oriented forwards to operate under a smaller noise distribution. As a result, the forward outputs become more representative of the noise-free output distribution and less reflective of potential output perturbations. Our findings show that this reversed adjustment reduces inference accuracy by 14.14\% compared to the standard NeFT configuration, substantiating the idea that less representative forward outputs (i.e., those bearing larger noise) can more effectively constrain robust convergence. By exposing the model to greater variability during training, OVF is better equipped to correct significant deviations in the backbone weights, thus ensuring resilience against 
nontrivial device variations. In contrast, using smaller $\sigma_d$ in the oriented forwards fails to constrain the network sufficiently, allowing subtle misalignments or noise sensitivities to persist into the final solution.

\textbf{Noise in Classifiers.}
Recall from Section~\ref{sec: IRS method} that noise is also injected into each classifier’s weights to capture the effects of device variation within the backbone and communicate these variations to the objective via representation snapshots. This design choice enables the backbone to benefit from negative feedback throughout training. In a subsequent experiment, we removed noise from the feedback classifiers during training, observing that the inference accuracy drops to 48.44\%, a significant decrease relative to the 65.81\% achieved under the standard NeFT configuration. These findings confirm the crucial role played by noisy classifiers in conveying variation information within NeFT training.

\textbf{Pretrained Models.}
A natural question is whether pretrained models can reduce the time and energy spent on training under NeFT. To explore this possibility, we utilize a vanilla-trained model (200 epochs, achieving 92.71\% accuracy at $\sigma_d=0$) as the initialized pretrained model for NeFT. Table~\ref{tab:pretrain} summarizes the noisy inference accuracy following various amounts of additional training (0, 50, 100, or 200 epochs) versus training from scratch (marked with $^*$). Although the pretrained model significantly shortens the initial training phase, its final accuracy under noise remains inferior to that of the model trained entirely with NeFT from scratch. 

In particular, OVF and IRS observe substantial gaps in accuracy when re-initialized from the pretrained model, even after 200 more epochs of NeFT-based training, compared to the $^*$ configuration. These results indicate that gradually integrating negative feedback constraints into the training process from its earliest stages enables the network to more effectively accommodate device variations, thus achieving higher accuracy. Consequently, training from scratch under NeFT appears to be the optimal practice for robust model convergence in the presence of substantial noise.
\vspace{-0.2in}

\begin{table}[t]
\centering
\caption{Comparison of NeFT performance with and without pretrained models. The baseline pretrained model (200 epochs) achieves 92.71\% accuracy at $\sigma_d=0$. Results under noise are shown after additional NeFT training (0, 50, 100, or 200 epochs), alongside training from scratch.}
\label{tab:pretrain}
\resizebox{\columnwidth}{!}{%
\begin{tabular}{cccccc}
\hline
\multirow{2}{*}{Methods} & \multicolumn{4}{c}{Pretrained} & From Scratch \\
 & +0 ep & +50 ep & +100 ep & +200 ep & +200ep \\ \hline
NeFT(OVF) & 10.75 & 55.97 & 57.15 & 56.05 & 63.50$^*$ \\
NeFT(IRS) & 10.75 & 55.07 & 57.34 & 56.94 & 65.81$^*$ \\ \hline
\end{tabular}%
}
\vspace{-0.2in}
\end{table}

\subsection{Training Cost Analysis}
\label{sec:train cost}

\begin{figure}[]
  \centering
  \includegraphics[scale=0.32]{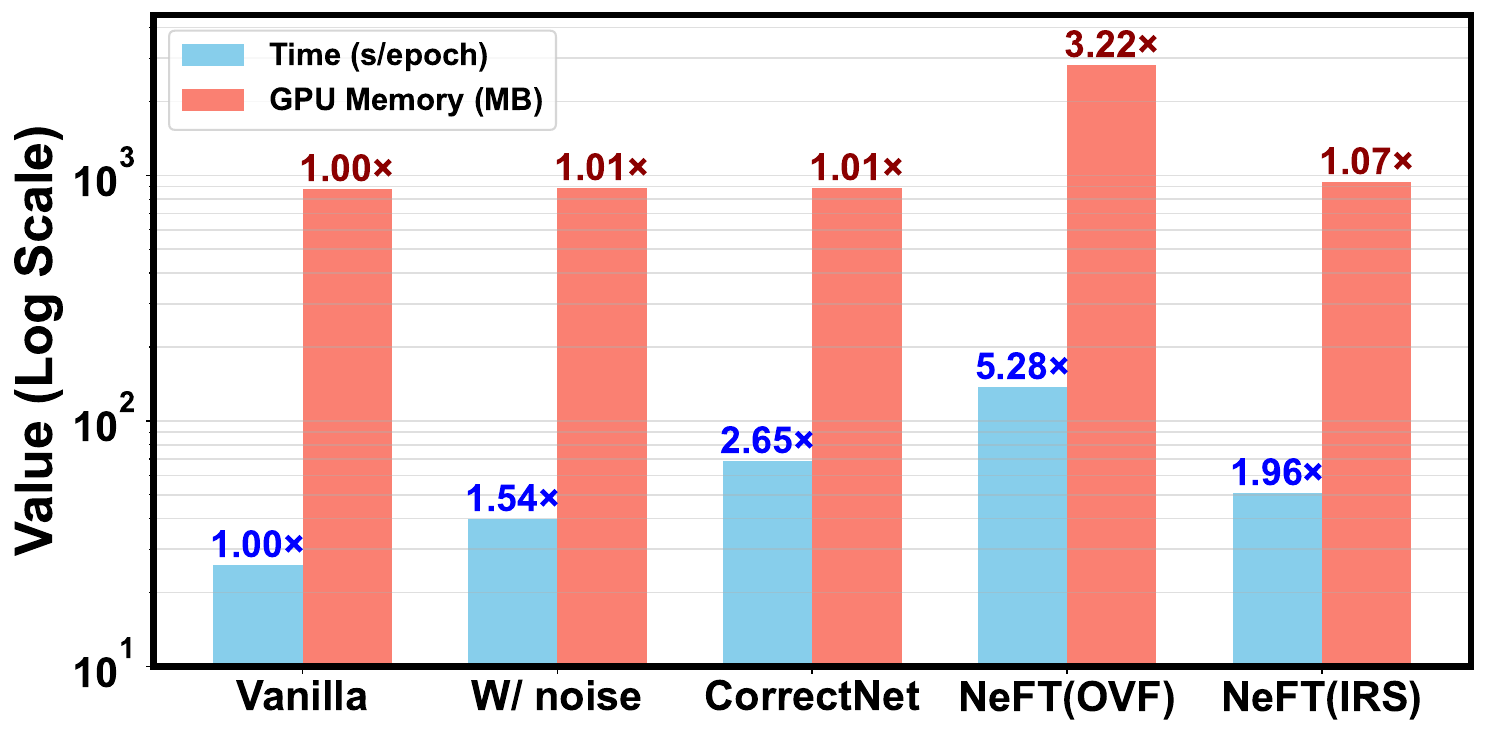}
  \caption{
Training cost comparison in terms of per-epoch runtime and peak GPU memory usage for different methods on the ResNet-18 architecture with the CIFAR-10 dataset. To facilitate visual comparison, training time values are linearly scaled to match the memory value range. The labels on top of each bar indicate the relative cost with respect to the baseline method (Vanilla).
}
\label{fig:training_cost}
\vspace{-0.2in}
\end{figure}

While NeFT significantly improves robustness against device variations, it introduces additional computational overhead during training. To quantify this cost, we measure the per-epoch training time and peak GPU memory usage for various methods under identical settings using ResNet-18 on the CIFAR-10 dataset. As shown in Figure~\ref{fig:training_cost}, OVF exhibits the highest training time, followed by IRS. Regarding GPU memory, OVF consumes 3.22$\times$ more memory due to feedback storage, whereas IRS remains lightweight at only 1.07$\times$ overhead.

Considering the trade-off between robustness and resource cost, we recommend using OVF for smaller models and datasets where higher feedback granularity benefits convergence, and IRS for larger-scale scenarios due to its better scalability and efficiency.

Despite the increased training cost, NeFT produces one-effort robust weight models for different accelerators, and models are directly compatible with existing CIM hardware. Furthermore, our method introduces \emph{no additional inference-time overhead}—no extra parameters, structural changes, or feedback components are retained after training. This makes NeFT highly suitable for deployment in low-latency and resource-constrained CIM applications.
It is important to note that this work focuses primarily on enhancing robustness, rather than optimizing training efficiency. In future work, we will investigate methods to reduce training costs, including sparse feedback scheduling, mixed-precision implementations, and hardware-aware gradient routing.

\section{Conclusion}\label{sec:conclusion}


In this study, we introduce Negative Feedback Training (NeFT) to enhance the robustness of NVCIM DNN accelerators, backed by a thorough theoretical analysis. We further propose two concrete implementations, oriented variational forward (OVF) and intermediate representation snapshot (IRS), which underscore NeFT’s generality and practicality in improving DNN robustness. Extensive experimental results across diverse architectures and datasets reveal that our method achieves notable improvements over state-of-the-art baselines, mitigating the impact of device variations, reducing uncertainty in model predictions, and stabilizing model convergence performance during training. Our NeFT method provides a potential solution for developing robust, reliable, and high-performance DNN accelerators.

\section{Acknowledgment}

This work is partially supported by the National Science Foundation (NSF) under Grants No. 2349538 and No. 2401544.

\bibliographystyle{IEEEtran}
\bibliography{Ref}

\clearpage
\begin{IEEEbiography}[{\includegraphics[width=1in,height=1.25in,clip,keepaspectratio]{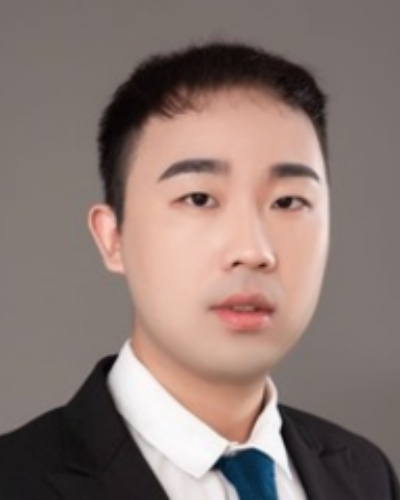}}]{Yifan Qin} 
received B.S. and M.S. degrees from Huazhong University of Science and Technology in 2017 and 2021. He is a Ph.D. student in the Department of Computer Science and Engineering at the University of Notre Dame, co-advised by Prof. Yiyu Shi and Prof. Sharon Hu. His research interests lie in software-hardware co-design of deep neural network accelerators, efficient AI and hardware.
\end{IEEEbiography}

\begin{IEEEbiography}[{\includegraphics[width=1in,height=1.25in,clip,keepaspectratio]{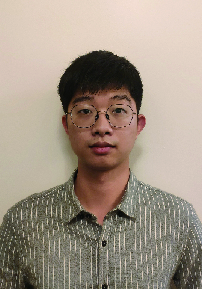}}]{Zheyu Yan} 
received a B.S. degree from Zhejiang University in 2019. He received his PhD degree from the Department of Computer Science and Engineering at the University of Notre Dame in 2024. His research interests lie in software-hardware co-design of deep neural network accelerators using emerging technologies, especially non-volatile memory-based compute-in-memory platforms.
\end{IEEEbiography}

\begin{IEEEbiography}[{\includegraphics[width=1in,height=1.25in,clip,keepaspectratio]{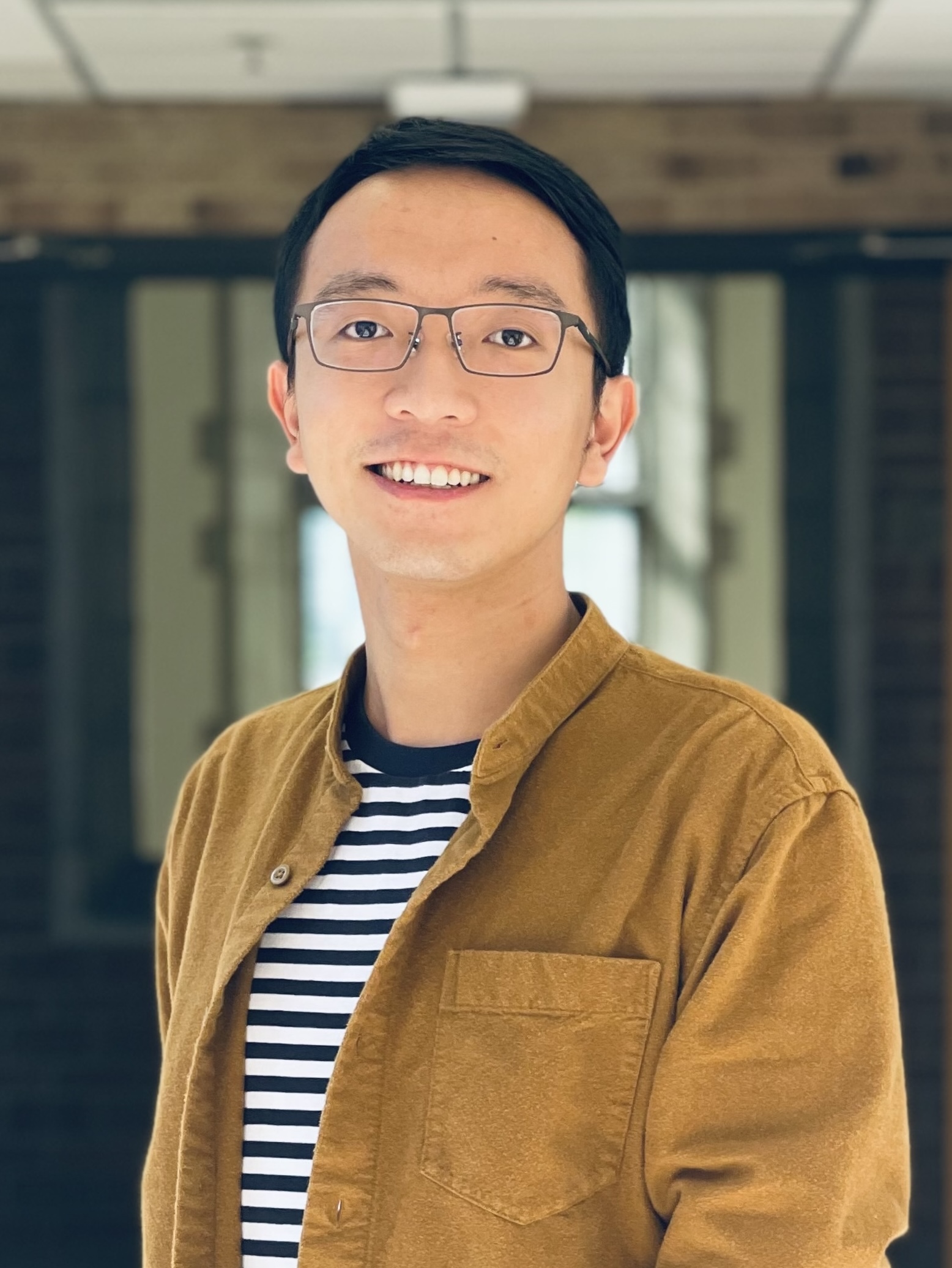}}]{Dailin Gan} received his B.S. in Statistics with First Class Honors from the University of Hong Kong, Hong Kong SAR, 2021. He is working towards a PhD in Statistics at the Department of Applied and Computational Mathematics and Statistics, University of Notre Dame, under the supervision of Prof. Jun Li. His current research interests lie in the development of statistical methods with application in single-cell omics data. 
\end{IEEEbiography}

\begin{IEEEbiography}[{\includegraphics[width=2.2cm]{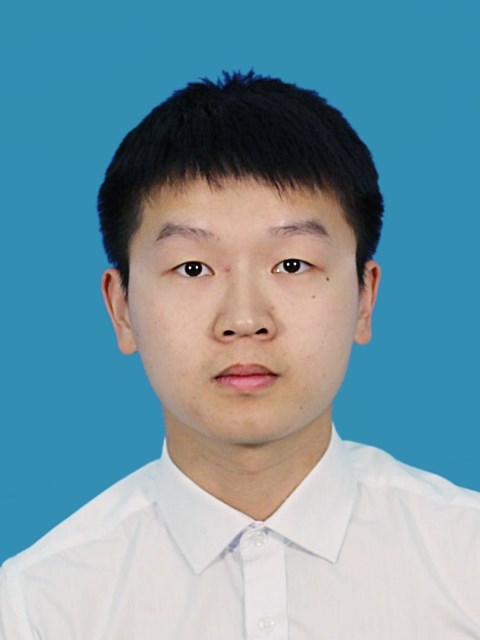}}]{Jun Xia}
received the B.S. degree from the Department of Computer Science and Technology, Hainan University in 2016, the M.E. degree from the Department of Computer Science and Technology, Jiangnan University in 2019, and the Ph.D. degree from the Department of Software Engineering, East China Normal University in 2023. He is currently a postdoc research fellow with the Department of Computer Science and Engineering, University of Notre Dame. His research interests are in the areas of  Federated Learning, and Responsible AI.
\end{IEEEbiography}

\begin{IEEEbiography}[{\includegraphics[width=1in,height=1.25in,clip,keepaspectratio]{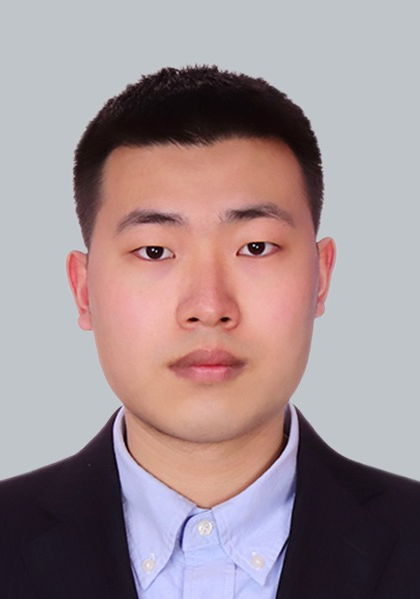}}]{Zixuan Pan}
received a bachelor’s degree in information engineering from Zhejiang University, China, in 2022. He is working toward a PhD at the University of Notre Dame under the supervision of Prof. Yiyu Shi. His current domains of interest include deep learning for biomedical applications, and computer vision.  
\end{IEEEbiography}

\begin{IEEEbiography}[{\includegraphics[width=1in,height=1.2in,clip,keepaspectratio]{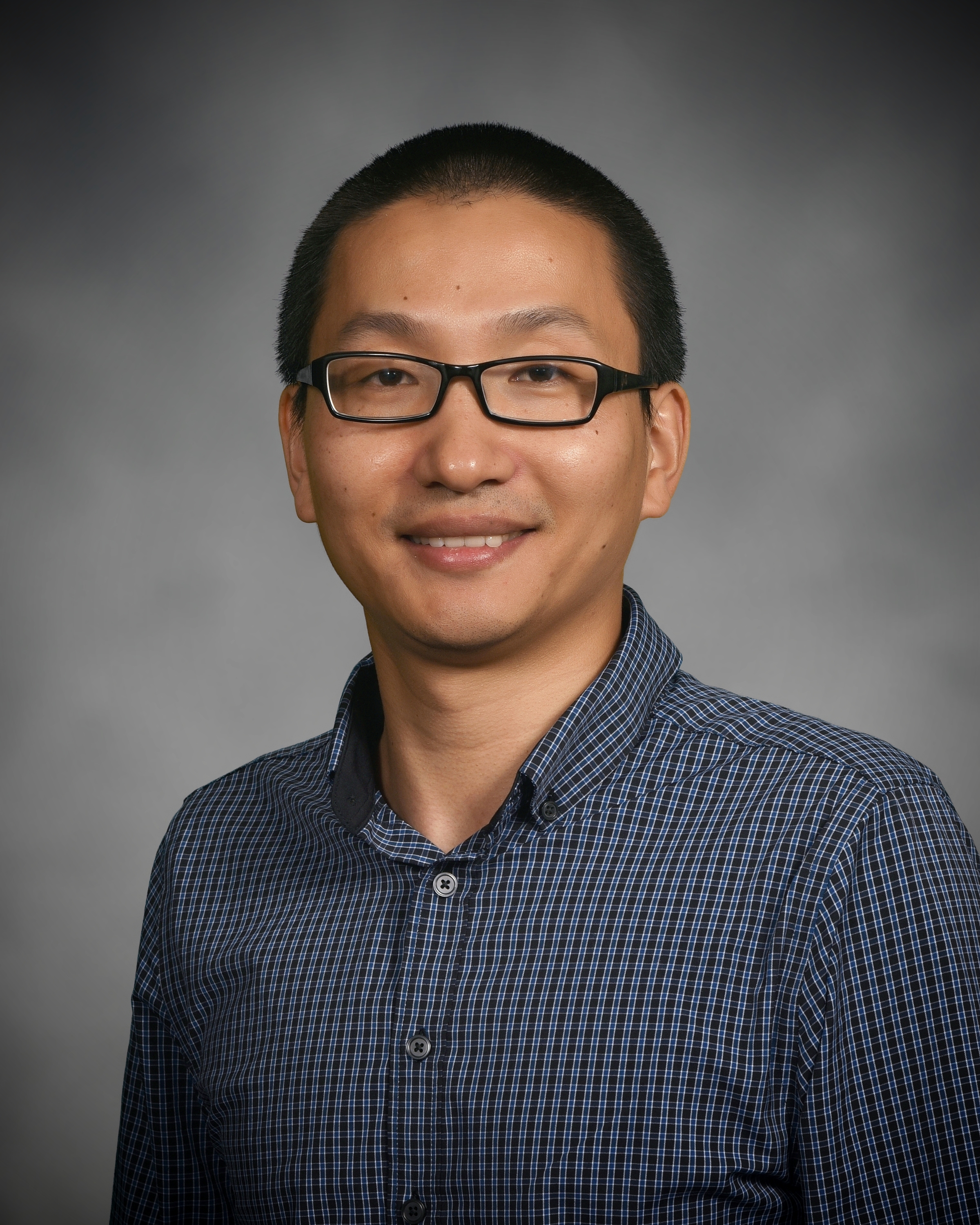}}]{Wujie Wen} is currently an Associate Professor in the Department of Computer Science at North Carolina State University (NCSU). He received his B.S. degree in electrical and computer engineering from Beijing Jiaotong University in 2006, Beijing, China, M.S. degree in communication engineering from Tsinghua University in 2010, Beijing China, and his Ph.D. degree in computer engineering from the University of Pittsburgh in 2015, Pittsburgh USA. Prior to joining the NCSU faculty, he was an assistant professor and then an associate professor in the Department of Electrical and Computer Engineering at Lehigh University. 

Dr. Wen received best paper nominations from all major EDA conferences and recently received the prestigious 2023 IEEE/ACM William J. McCalla ICCAD Best Paper Award at the 42nd ACM/IEEE Conference on Computer-Aided Design (ICCAD). He has published extensively on CSRankings conference, including DAC, ICCAD, MICRO, HPCA, IEEE Security and Privacy (Oakland), USENIX Security, NeurIPS, ICML, CVPR, ICCV, ECCV, AAAI etc. He served as general chair and program chair of the IEEE Computer Society Annual Symposium on VLSI (ISVLSI) 2018 and 2019, respectively. He served/serves as an Associate Editor of Neurocomputing, IEEE Circuit and Systems Magazine and ACM Transactions on Design Automation of Electronic Systems (ACM TODAES). He is also a recipient of the NSF Faculty Early Career Award.
\end{IEEEbiography}

\begin{IEEEbiography}[{\includegraphics[width=1in,height=1.2in,clip,keepaspectratio]{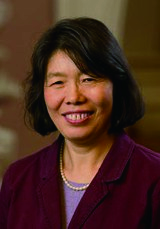}}]{Xiaobo Sharon Hu} 
(S'85-M'89-SM'02-F'16) received her B.S. degree from Tianjin University, China, in 1982, M.S. from Polytechnic Institute of New York in 1984, and Ph.D. from Purdue University, West Lafayette, Indiana in 1989. She is a Professor in the department of Computer Science and Engineering at the University of Notre Dame. Her research interests include energy/reliability-aware system design, circuit and architecture design with emerging technologies, real-time embedded systems, and hardware-software co-design. She has published more than 450 papers in these areas. 

Some of X. Sharon Hu's recognitions include the Best Paper Award from the Design Automation Conference, International Conference on Computer-Aided Design, and the International Symposium on Low Power Electronics and Design, as well as the NSF Career award. She is the Editor-in-Chief of ACM Transactions on Design Automation of Electronic Systems and also served as Associate Editor for IEEE Transactions on CAD, IEEE Transactions on VLSI,  ACM Transactions on Embedded Computing, etc.  She served as the General chair and Technical Program Chair of Design Automation Conference (DAC), IEEE Real-time Systems Symposium, etc. X. Sharon Hu is a Fellow of the ACM and a Fellow of the IEEE.
\end{IEEEbiography}

\begin{IEEEbiography}[{\includegraphics[width=1in,height=1.2in,clip,keepaspectratio]{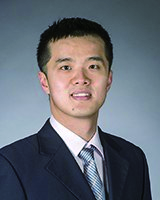}}]{Yiyu Shi} 
received the B.S. degree (Hons.) in electronic engineering from Tsinghua University, Beijing, China, in 2005, and the M.S. and Ph.D. degrees in electrical engineering from the University of California at Los Angeles, Los Angeles, CA, USA, in 2007 and 2009, respectively. He is currently a Professor with the Departments of Computer Science and Engineering and Electrical Engineering, University of Notre Dame, Notre Dame, IN, USA. His current research interests include 3-D integrated circuits, hardware security, and renewable energy applications. 

Prof. Shi was a recipient of several best paper nominations in top conferences, Facebook Research Award, IBM Invention Achievement Award, Japan Society for the Promotion of Science (JSPS) Faculty Invitation Fellowship, Humboldt Research Fellowship, IEEE St. Louis Section Outstanding Educator Award, Academy of Science (St. Louis) Innovation Award, Missouri S\&T Faculty Excellence Award, NSF CAREER Award, IEEE Region 5 Outstanding Individual Achievement Award, the Air Force Summer Faculty Fellowship, and IEEE Computer Society Mid-Career Research Achievement Award. He has served on the technical program committee of many international conferences.
\end{IEEEbiography}

\vfill

\end{document}